\documentclass[sigconf,nonacm]{acmart}

\AtBeginDocument{%
  \providecommand\BibTeX{{%
    \normalfont B\kern-0.5em{\scshape i\kern-0.25em b}\kern-0.8em\TeX}}}

\copyrightyear{2022}
\acmYear{2022}

\begin{document}

\title[Automatic Detection and Rectification of Paper Receipts on Smartphones]{Automatic Detection and Rectification of Paper Receipts on Smartphones}

\author{Edward Whittaker}
\author{Masashi Tanaka}
\affiliation{%
  \institution{Best Path Research Inc.}
  \streetaddress{Shintomi, Chuo-ku}
  \city{Tokyo}
  \country{Japan}
  \postcode{104-0041}
}
\email{ed@bestpathresearch.com} \email{tanaka@bestpathresearch.com}

\author{Ikuo Kitagishi}
\affiliation{%
  \institution{Money Forward Inc.}
  \streetaddress{Shibaura, Minato-ku}
  \city{Tokyo}
  \country{Japan}}
  \postcode{108-0023}
\email{kitagishi.ikuo@moneyforward.co.jp}

\renewcommand{\shortauthors}{Whittaker, Kitagishi, et al.}

\begin{abstract}
We describe the development of a real-time smartphone app that allows the user to digitize paper receipts in a novel way by ``waving'' their phone over the receipts and letting the app automatically detect and rectify the receipts for subsequent text recognition. 

We show that traditional computer vision algorithms for edge and corner detection do not robustly detect the non-linear and discontinuous edges and corners of a typical paper receipt in real-world settings. This is particularly the case when the colors of the receipt and background are similar, or where other interfering rectangular objects are present. Inaccurate detection of a receipt's corner positions then results in distorted images when using an affine projective transformation to rectify the perspective.

We propose an innovative solution to receipt corner detection by treating each of the four corners as a unique ``object'', and training a Single Shot Detection MobileNet object detection model. We use a small amount of real data and a large amount of automatically generated synthetic data that is designed to be similar to real-world imaging scenarios. This data is created by placing randomly selected images of real receipts on randomly selected images of real backgrounds. Random projective transformations are then applied to simulate a user taking a picture of a real receipt with the camera placed at different inclinations, rotations and rolls relative to the receipt and background. 

We show that our proposed method robustly detects the four corners of a receipt, giving a receipt detection accuracy of 85.3\% on real-world data, compared to only 36.9\% with a traditional edge detection-based approach. Our method works even when the color of the receipt is virtually indistinguishable from the background.

Moreover, our method is trained to detect only the corners of the central target receipt and implicitly learns to ignore other receipts, and other rectangular objects. Combining real-world and synthetic data allows us to train an even better model. These factors are a major advantage over traditional edge detection-based approaches, allowing us to deliver a much better experience to the user.

Our approach is currently being integrated into the ``Money Forward ME'' app which has over 12 million users in Japan.
\end{abstract}




\begin{teaserfigure}
  \includegraphics[width=\textwidth]{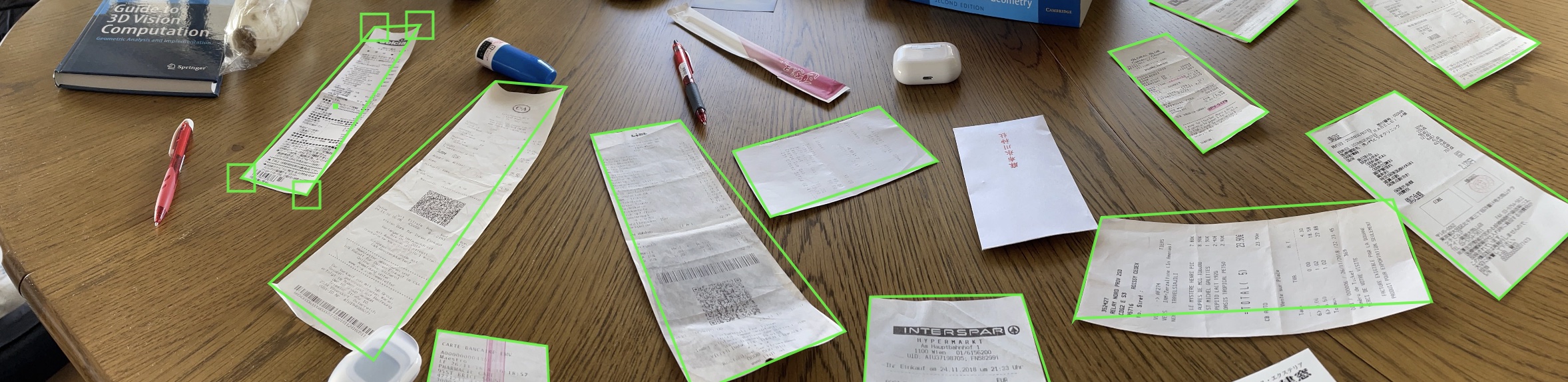}
\caption{Haphazard layout of receipts (highlighted) to be detected and rectified by ``waving'' a phone over them.}
  \Description{}
  \label{fig:teaser}
\end{teaserfigure}

\maketitle

\section{Introduction}

Despite a recent trend worldwide of payments moving from physical to digital, paper receipts are often still issued upon payment in physical stores. Such paper receipts serve several functions, for example as proof of purchase against claims of theft, as proof of expenses to be shown to an employer or to the tax authorities, as well as proof of purchase when returning an item. Indeed the paper receipt often contains much more information than can be acquired through purely ``back-end payment processors'' (e.g. credit card companies). Such information might include the number and identity of the individual items purchased, the time and location of purchase, and whether coupons and/or multiple payment methods were used. Consequently, we might conclude that the ubiquitous paper receipt is unlikely to disappear any time soon. 

There already exist several smartphone apps which can capture and digitise paper receipts such as Apple Notes\footnote{https://www.icloud.com/notes/}, Expensify\footnote{https://www.expensify.com/}, and Zoho \footnote{https://www.zoho.com/expense/}. Another such app, which has over 12 million users in Japan and processes over 4 million receipt images on average per month, is {\it Money Forward ME}\footnote{https://apps.apple.com/jp/app/wu-liao-jia-ji-bu-manefowado/id594145971}.

\begin{figure}[h]
  \centering
  \includegraphics[width=3.5cm]{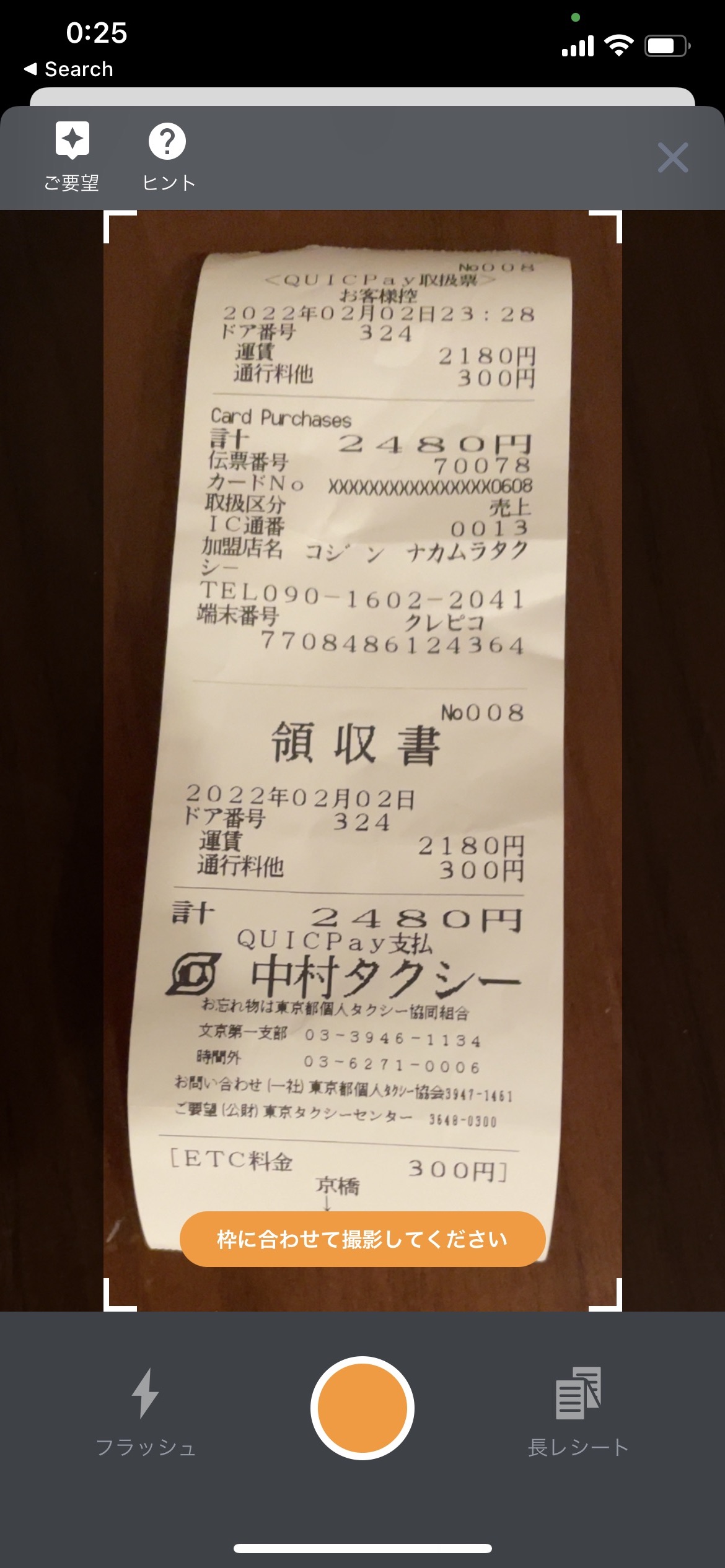}
  \caption{Screenshot of the Money Forward ME app's receipt acquisition screen showing the bounding rectangle in which a user should position a paper receipt.}
  \Description{}
  \label{Figure_MoneyForward_ME}
\end{figure}

Most such apps require a somewhat tedious and error-prone alignment of the paper receipt inside a bounding rectangle, or inside virtual cross-hairs shown on-screen, like the screenshot of the Money Forward ME app shown in Figure~\ref{Figure_MoneyForward_ME}. The simple act of pressing the button to take a photo can shift the position of the camera and blur the acquired image. Moreover, the user may need to physically stand up to take an overhead photo of a receipt placed on a table if they are unable to hold the receipt directly in front of their phone. Therefore, a method to automatically detect and rectify the image of a receipt when it appears in the field of view is desirable, both from the point of view of simplifying the process for the user, and the point of view of improving downstream processing tasks, such as optical character recognition and named entity extraction.

In this paper, we propose a user interface whereby the user ``waves'' their smartphone over a collection of receipts (e.g. laid out on a flat surface, or held by hand) and possibly with a background containing other confusable objects, as shown in Figure\ref{fig:teaser}. We leave it up to the app to detect, rectify, and store each receipt image automatically.

An important part of our imaging pipeline is the image rectification step. We use the four-point perspective transform\footnote{\texttt{https://docs.opencv.org/4.x/da/d54/group\_\_imgproc\_\_transform.html\\\#ga20f62aa3235d869c9956436c870893ae}\\\texttt{https://www.pyimagesearch.com/2014/08/25/\\4-point-opencv-getperspective-transform-example/}} which uses the correspondences between four points (in our case, the four corners of a receipt) to compute the transformation matrix. Corners on real receipts are characterized by edges that are not straight, do not converge at right angles and may not even be continuous lines. Some corners might even be invisible, yet a human could reliably infer where they would be using the text on the receipt, the location of other corners and the location of partially visible edges as a guide.

Our task is similar but different to the ICDAR 2019 Robust Reading Challenge on Scanned Receipts OCR and Information Extraction (SROIE)\footnote{\texttt{https://rrc.cvc.uab.es/?ch=13}} \cite{DBLP:conf/icdar/2019}\cite{Huang2019ICDAR2019CO} where receipts were captured in rectangular form from a top-down, {\it bird's-eye-view} perspective using flatbed scanners. The captured receipts do, however, share other characteristics with ours, such as deteriorated paper conditions, poor ink and print quality, and distorted images with low resolution of the target receipt region, varied receipt sizes and aspect ratios. Unfortunately, we are currently unable to release our annotated dataset for privacy reasons. However, we hope to do so in the near future. 

Edge and corner detection have been studied extensively in the Computer Vision community for many decades. Implementations of many algorithms are available that have been optimised for speed and power on both iOS\footnote{\texttt{https://developer.apple.com/documentation/vision}} and Android\footnote{\texttt{https://developers.google.com/vision}\\ \texttt{https://developers.google.com/ml-kit}}. However, it is well-known that these algorithms require many hand-tuned thresholds and parameters to be set, which makes them brittle in the face of real-world (hereafter shortened to: real) images, which often have highly variable contrast, brightness and blurriness.

In Section~\ref{Baseline_Approach} we describe our attempts at using traditional computer vision algorithms to perform edge detection on receipts, and the downsides of such an approach. Based on these observations we hypothesised that a reasonably straightforward object detection model might be trained to detect the four unique corners of a receipt if enough realistic training examples were available. We also hypothesised that the model might even learn to differentiate the corners of receipts as distinct from the corners of other rectangular objects. Indeed, we take this one step further during our experiments and train the model to detect only the four corners of a central target receipt, while ignoring the corners of other ``interfering'' receipts that may also happen to be in view. We describe our proposed approach in Section~\ref{Proposed_Approach}. Note that training a model to detect whole receipt objects would produce bounding boxes that are not sufficiently flush with the borders of the receipt unless the receipt was captured from directly overhead so such a model is unsuitable for our task. On the other hand, a segmentation model that classified pixels as belonging to the target receipt, might work but would still not tell us where the corners are. Moreover, at the time of writing such segmentation models are still too slow to deploy on smartphones if we want to process images at a high frame-rate.

For our proposed approach, we take advantage of recent progress in low-latency object detection models, such as the Single Shot Detection (SSD) MobileNet model\cite{DBLP:conf/cvpr/HuangRSZKFFWSG017}, which have been optimised for running on smartphones, and focus instead on the task of data generation. 

Since collecting and manually rectifying a large number of real receipt images would be inefficient and expensive, we take a corpus of scanned and closely cropped receipts (i.e.~where receipt edges match the image boundaries and the corners are at right angles), and apply random projective transformations to simulate the variety of orientations in which a user might take a photo of a receipt with their phone. The model should thus learn to recognise robustly the many ways in which the corners of receipts might actually appear in reality.

Once the four corners of the target receipt have been correctly identified, we can then apply an affine transformation to rectify the receipt. Note that we do not train a model to learn the transformation itself, or to perform non-linear image rectification, as has been investigated elsewhere in the literature\cite{Ma_2018_CVPR}\cite{DBLP:journals/corr/YouMSBI16}. 

We describe our baseline approach in Section~\ref{Baseline_Approach} followed by our proposed approach in Section~\ref{Proposed_Approach}. A description of the real data and the synthetic data that we use in our experiments is given in Section~\ref{Data} and our experimental work is described in Section~\ref{Experiments}. Results are given in Section~\ref{Results} and a discussion of them in Section~\ref{Discussion} followed by our conclusions and ideas for future work in Section~\ref{Conclusions}.

\section{Baseline Approach}\label{Baseline_Approach}

We use only traditional computer vision techniques to derive a baseline accuracy on receipt corner detection. We employ the approach described in pp.131-140 of \cite{Pyimagesearch_OCR_Book} for which there is also an implementation publicly available\footnote{\texttt{https://www.pyimagesearch.com/2014/09/01/\\build-kick-ass-mobile-document-scanner-just-5-minutes/}}. A series of image processing functions are applied to estimate where the edges of a receipt are. These edges are then used to determine the location of the four corners of the receipt. Based on knowledge of the typical aspect ratio of a receipt we can guess the likely identities of each corner. The four corners are then used to compute the transformation matrix which can be used to transform the captured image to a rectangular image in the so-called ``bird's-eye-view''.

\begin{figure}[h]
  \centering
  \includegraphics[width=2.7cm]{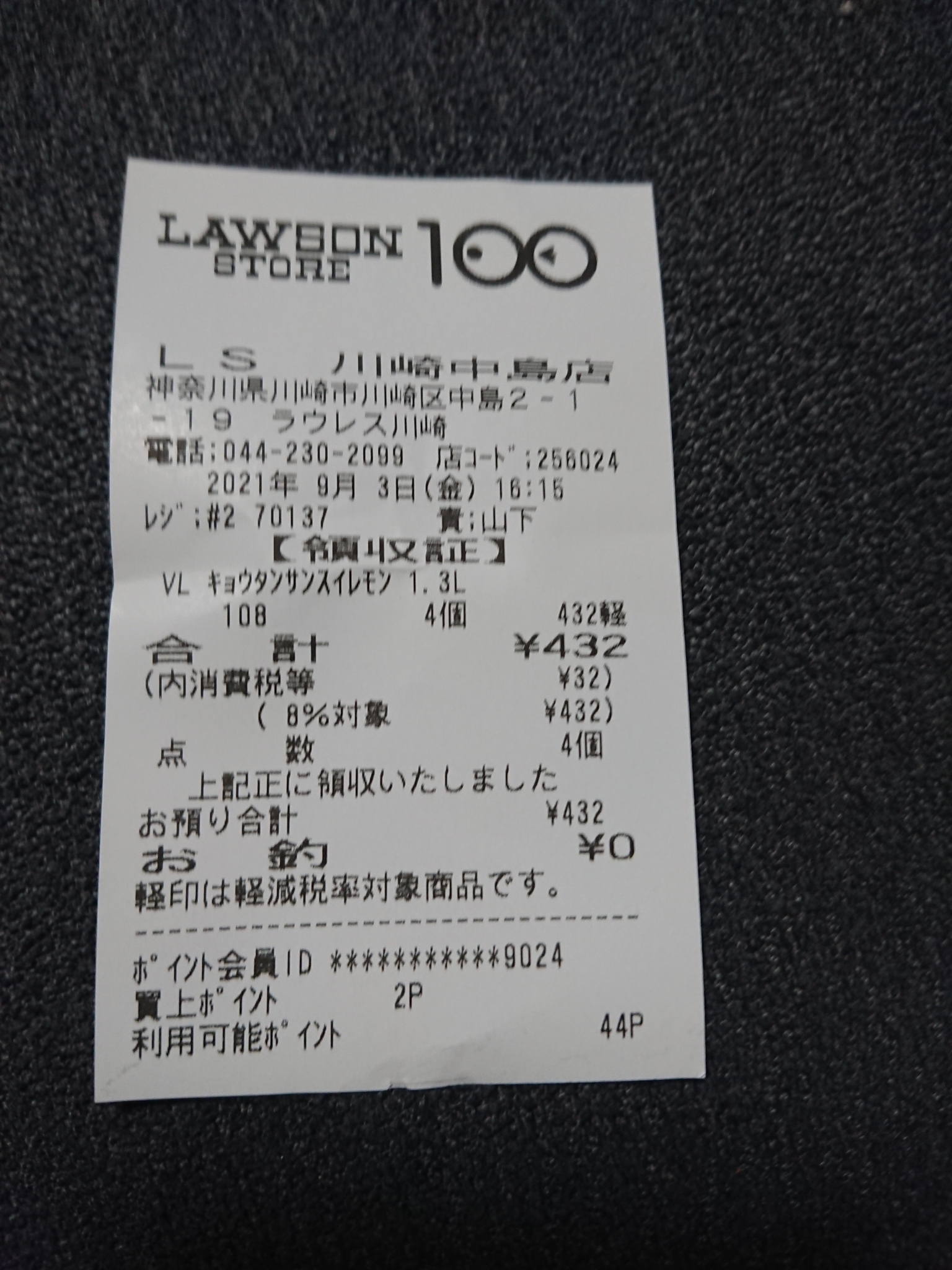}
  \includegraphics[width=2.7cm]{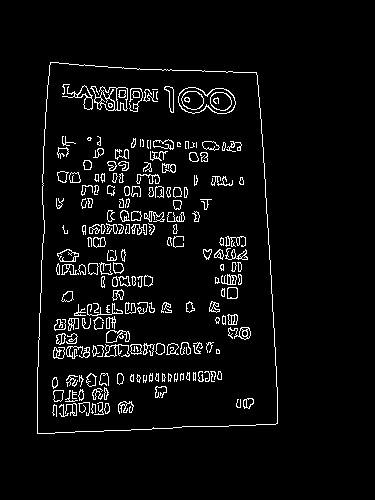}
  \includegraphics[width=2.7cm]{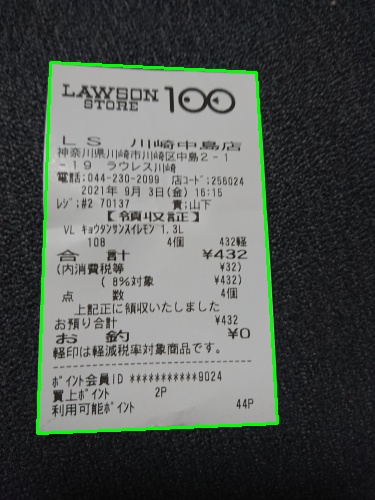}
      \caption{Example using the baseline edge detection method on a high contrast example image (left), after Gaussian blurring and edge detection (center), and showing the detected quadrilateral region with the largest area highlighted in bright green (right).}
  \Description{}
  \label{edge_detection_example}
\end{figure}

More precisely, we first convert the color image of our receipt to a monochrome one, then apply Gaussian blurring (with a tunable, odd-numbered kernel size) to reduce high-frequency image noise. We then apply the Canny edge detector\cite{Canny86} which finds all the lines with a gradient above a certain tunable threshold (of value between 0 and 500). The output of the Canny edge detector is a binary image, with 1 for detected edges and 0 for others. The image after Canny edge detection is shown in the center of Figure~\ref{edge_detection_example}. Note that the Canny edge detector will find {\it all} the lines in the image, not just those associated with a receipt edge.

We then use the algorithm described in~\cite{Suzuki85} to find all the contours in the image. The result of this operation is a set of polygons which we then sort by their area. We select the polygon with the largest area as being the target receipt region.

Next we use the Douglas-Peucker Algorithm~\cite{doi:10.3138/FM57-6770-U75U-7727} to approximate the polygon as a quadrilateral. In this algorithm, there is a parameter called epsilon, used to determine the approximation accuracy, which is a tunable fraction of the length of the perimeter.

We then check whether the output consists of four edges and if it does, we output the four vertices of these contours as the corners of the receipt. The detected receipt is shown on the right in Figure~\ref{edge_detection_example}. Finally, we compute the projective transformation using the four detected corner positions\footnote{\texttt{https://docs.opencv.org/4.x/da/d54/group\_\_imgproc\_\_transform.html\\\#ga20f62aa3235d869c9956436c870893ae}\\\texttt{https://www.pyimagesearch.com/2014/08/25/\\4-point-opencv-getperspective-transform-example/}}.

All these image processing functions are available in OpenCV\cite{opencv_library}. In Section~\ref{Experiments}, we explain how the parameters are optimised on a held-out validation set of real receipt images.

\section{Proposed Approach}\label{Proposed_Approach}

The traditional computer vision approach to edge detection, described in the previous section, was found to be very sensitive to lighting, shadows, and colors, particularly when the contrast between the receipt color and the background color, against which it is captured, is low. Moreover, such traditional methods do not distinguish between the edges of a receipt and the edges of other rectangular objects that might also be in view, such as text, books, letters and phones. 

Recall that our objective in this work is to identify and locate the four corners of a single target receipt which we assume are present in the image. (We exclude receipts with missing, or occluded, corners from the current investigation.)

It has been shown that CNN-based deep neural network object detectors learn to represent simple features, such as edges and corners, in the lower layers and learn progressively more complex features, such as facial features or vehicle features, in the higher layers\cite{zeiler2014visualizing}. Our proposed approach treats each of the four corners as a different unique ``object''.
We hypothesise that by training the model on a large amount of varied data, the model will learn to distinguish, say, the upper-left hand corner, from the other three corners. We believe this is not an unreasonable assumption given that each corner is likely to have very different positional distributions of (receipt) edges and content (e.g. receipt text vs. homogeneous background).

While using such a model might seem excessive for capturing what is essentially a simple geometric shape, we argue that there is actually a very large variation in the visual appearance of real corners, 
and that a neural network is well-suited to capturing such variation. With suitable training data we should even be able to detect corners that have curved and non-continuous edges, something that traditional methods also fail at. 

We define the center of each bounding box to be the exact corner whose position we want to determine. This makes the computation of corner locations trivial at inference-time once the object bounding boxes have been output. In addition, in order for the model to learn as much useful information as possible, we maximize the size of each bounding box while enforcing this constraint, increasing the size of the bounding box up to the closest edge of the training image while keeping the bounding box square i.e. all sides the same length.

We believe this problem formulation is a novel contribution that has potential application outside of receipt detection, for example in licence-plate or street-sign detection.

\section{Data}\label{Data}

For testing purposes, in this work we only use real images of receipts captured by users using a smartphone. However, we also use real data for training the models in our proposed approach, and also for hyper-parameter optimisation in both the proposed and baseline approaches. This real data is described in Section~\ref{RealData}.

In addition, we create {\it synthetic data} for training purposes only, which aims to simulate the kinds of images of receipts seen in the real data. How this synthetic data is created is described in detail in Section~\ref{syntheticData}.

We are currently unable to release the dataset used in these experiments for data privacy reasons, but hope to do so in future. In the meantime, below we describe how our results may be reproduced given a similar dataset of receipt images, such as those available in the SROIE data\footnote{\texttt{https://drive.google.com/drive/folders/\\1ShItNWXyiY1tFDM5W02bceHuJjyeeJl2}} \cite{Huang2019ICDAR2019CO} and trained using the publicly available Tensorflow object detection API training scripts\footnote{\texttt{https://github.com/tensorflow/models/tree/master/\\research/object\_detection}} for the SSD MobileNet model described in~\cite{DBLP:conf/cvpr/HuangRSZKFFWSG017}. This model assumes an input image size of 300x300 pixels so it should be noted that object detection is therefore performed at a significantly lower resolution than the images we prepare.

\subsection{Real data}\label{RealData}

\begin{figure}[h]
  \centering
  \includegraphics[width=2cm]{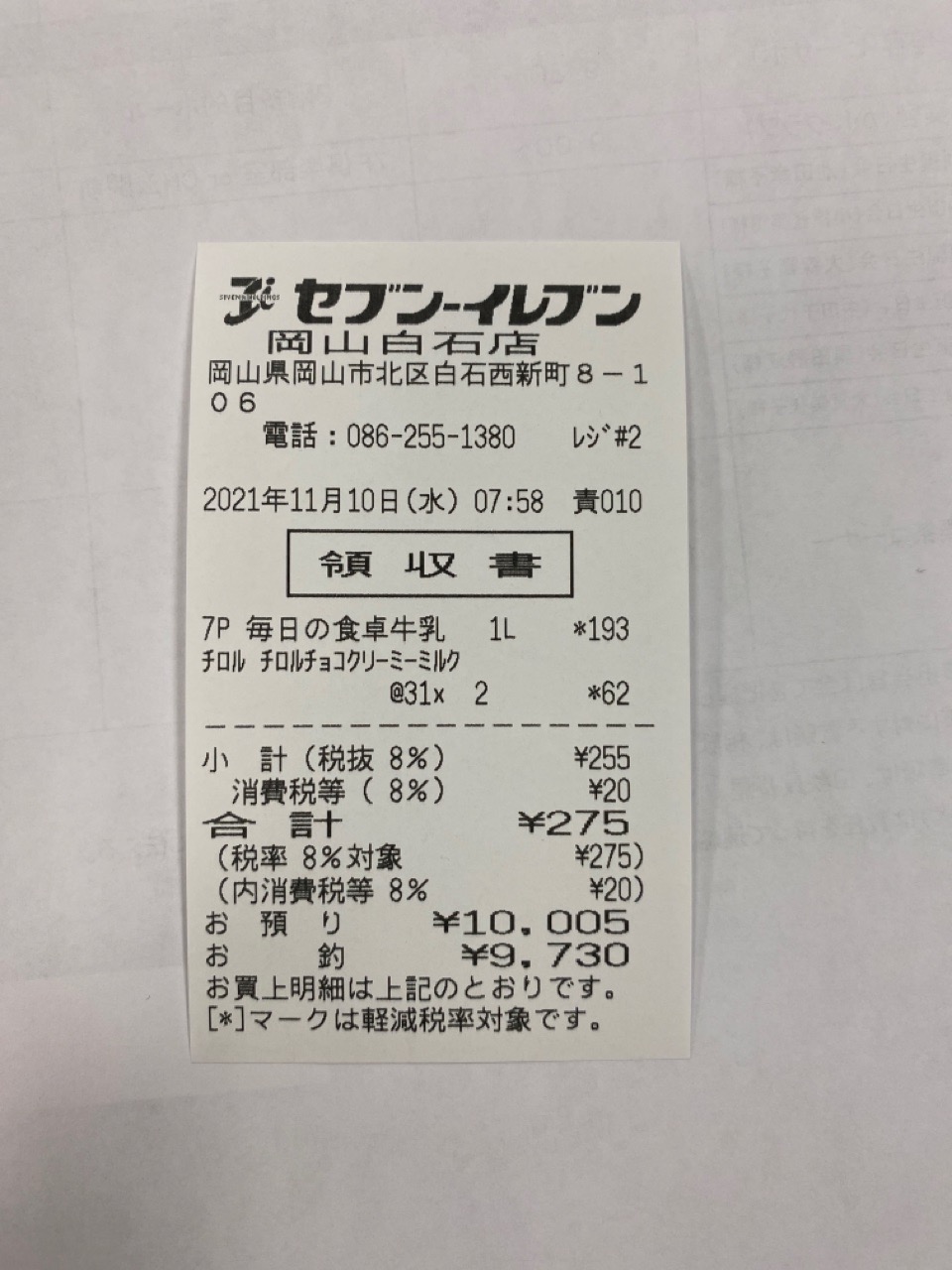}
  \includegraphics[width=2cm]{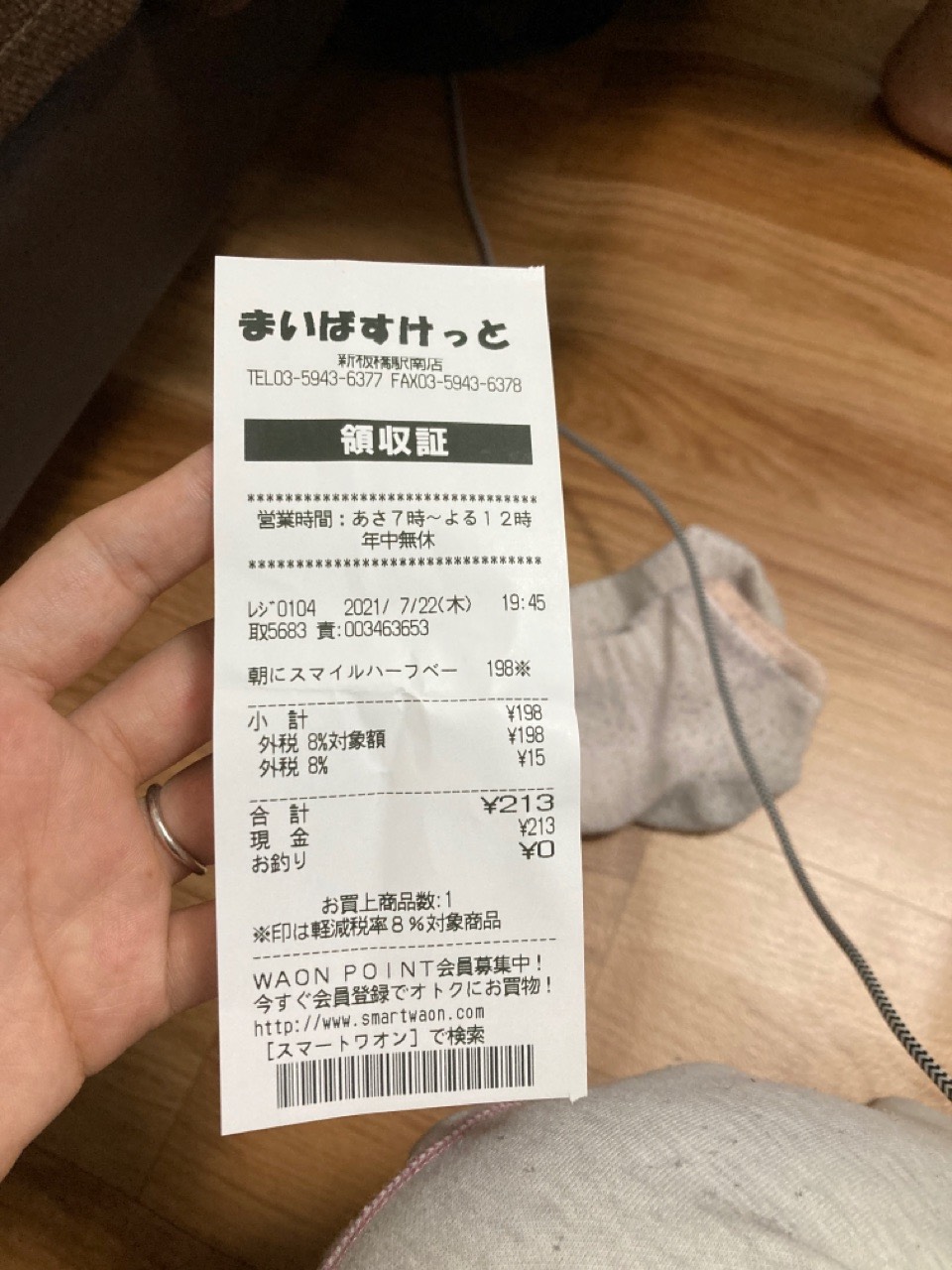}
  \includegraphics[width=2cm]{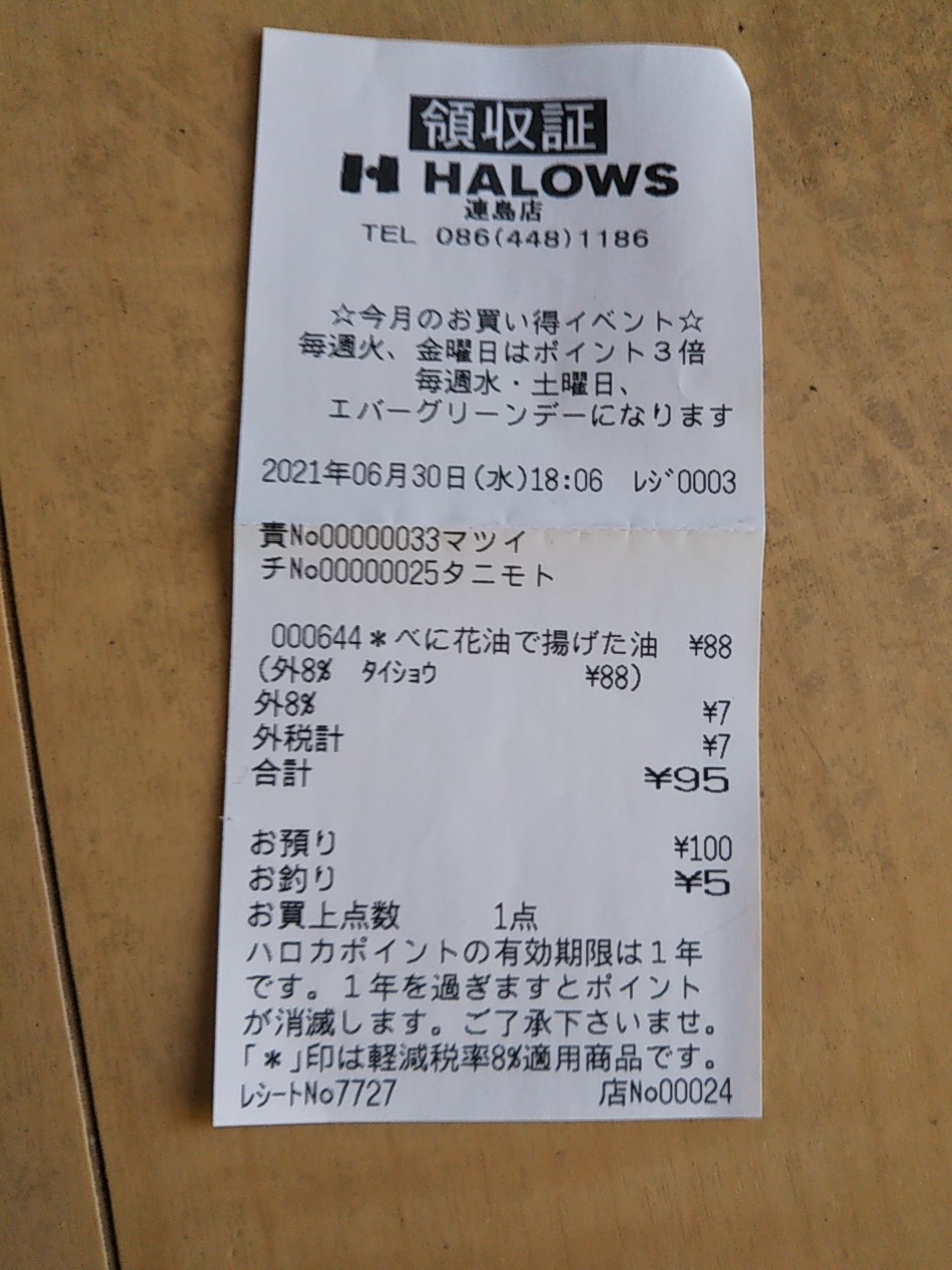}
  \includegraphics[width=2cm]{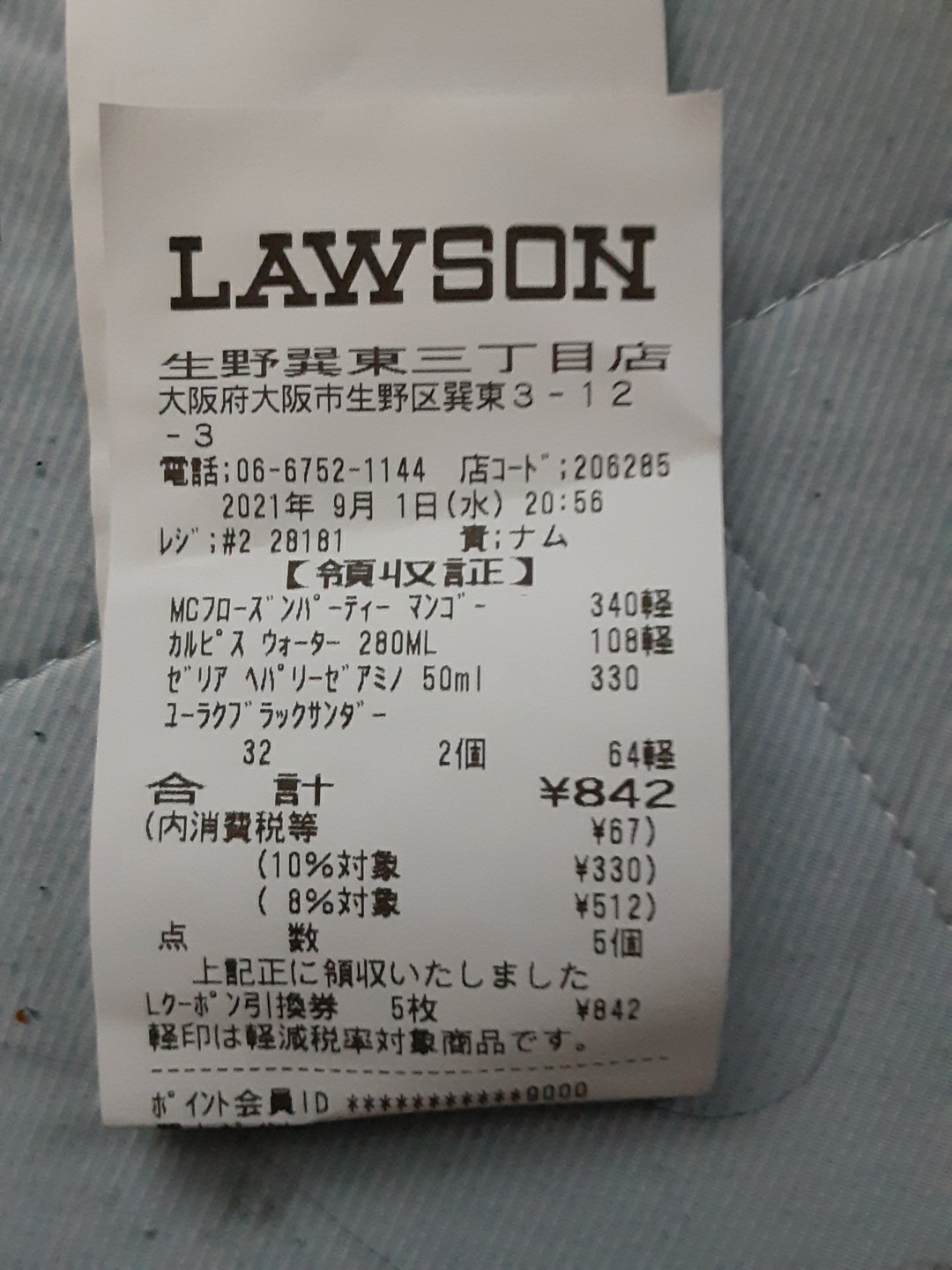}
  \includegraphics[width=2cm]{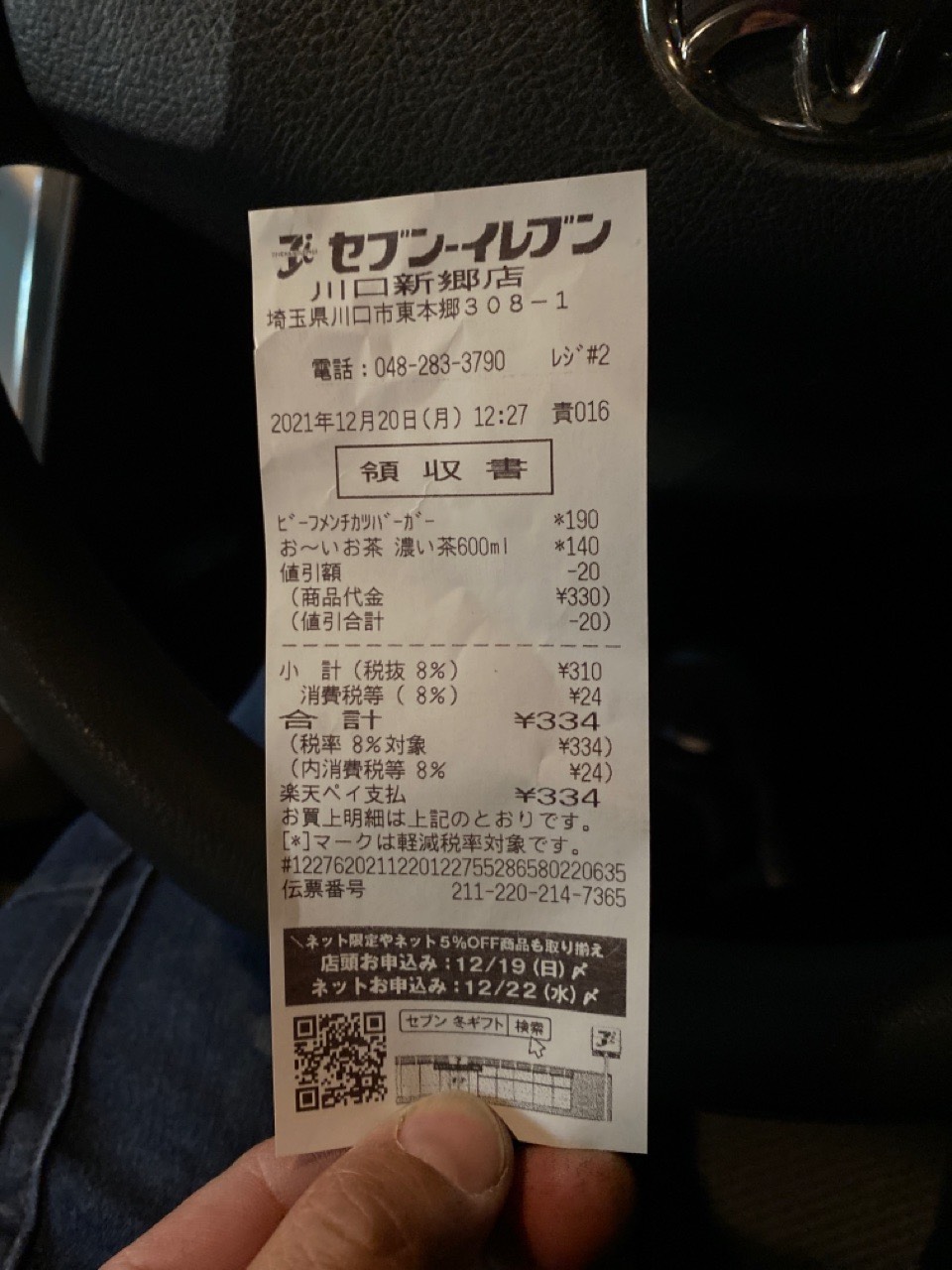}
  \includegraphics[width=2cm]{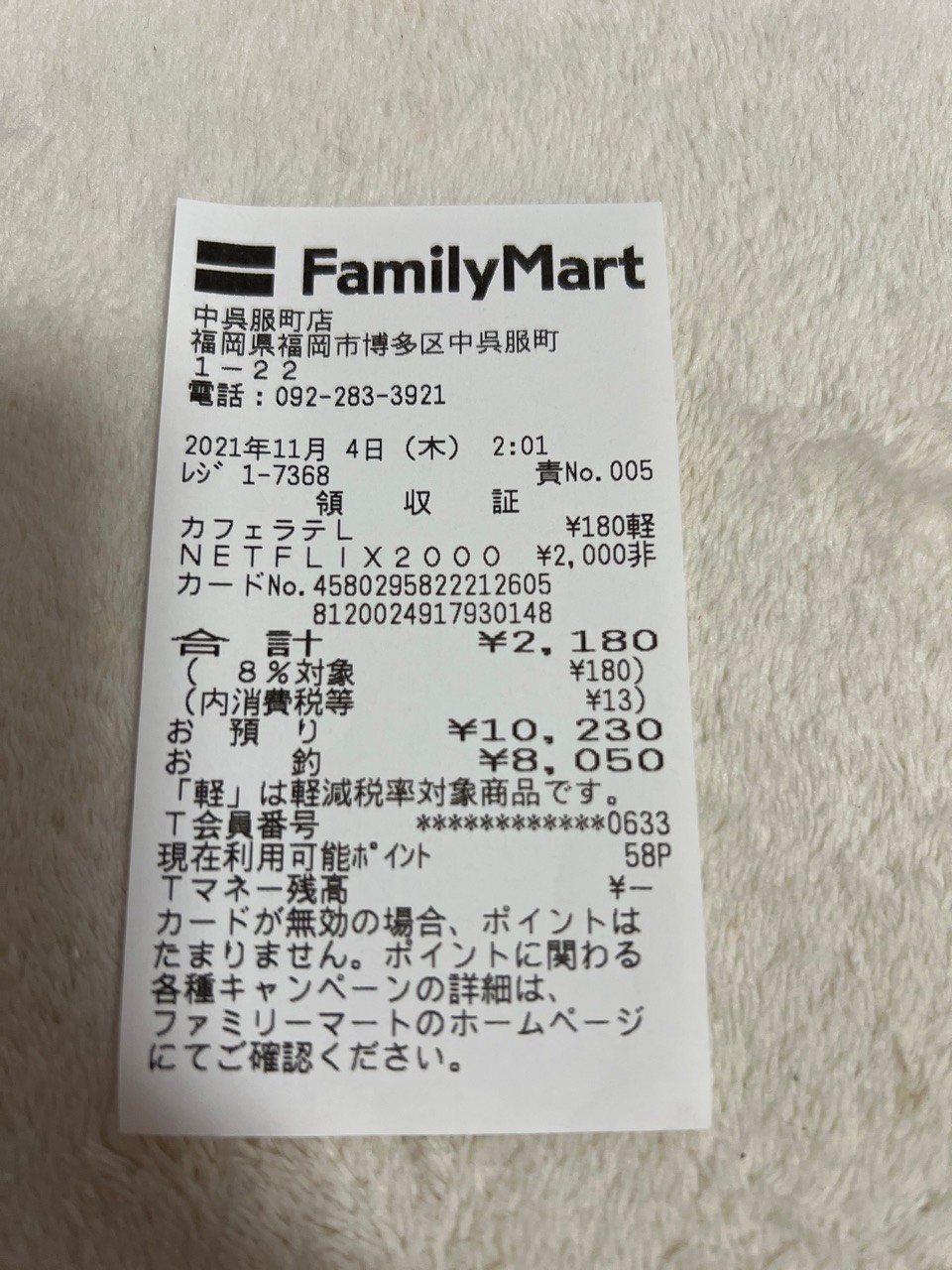}
  \includegraphics[width=2cm]{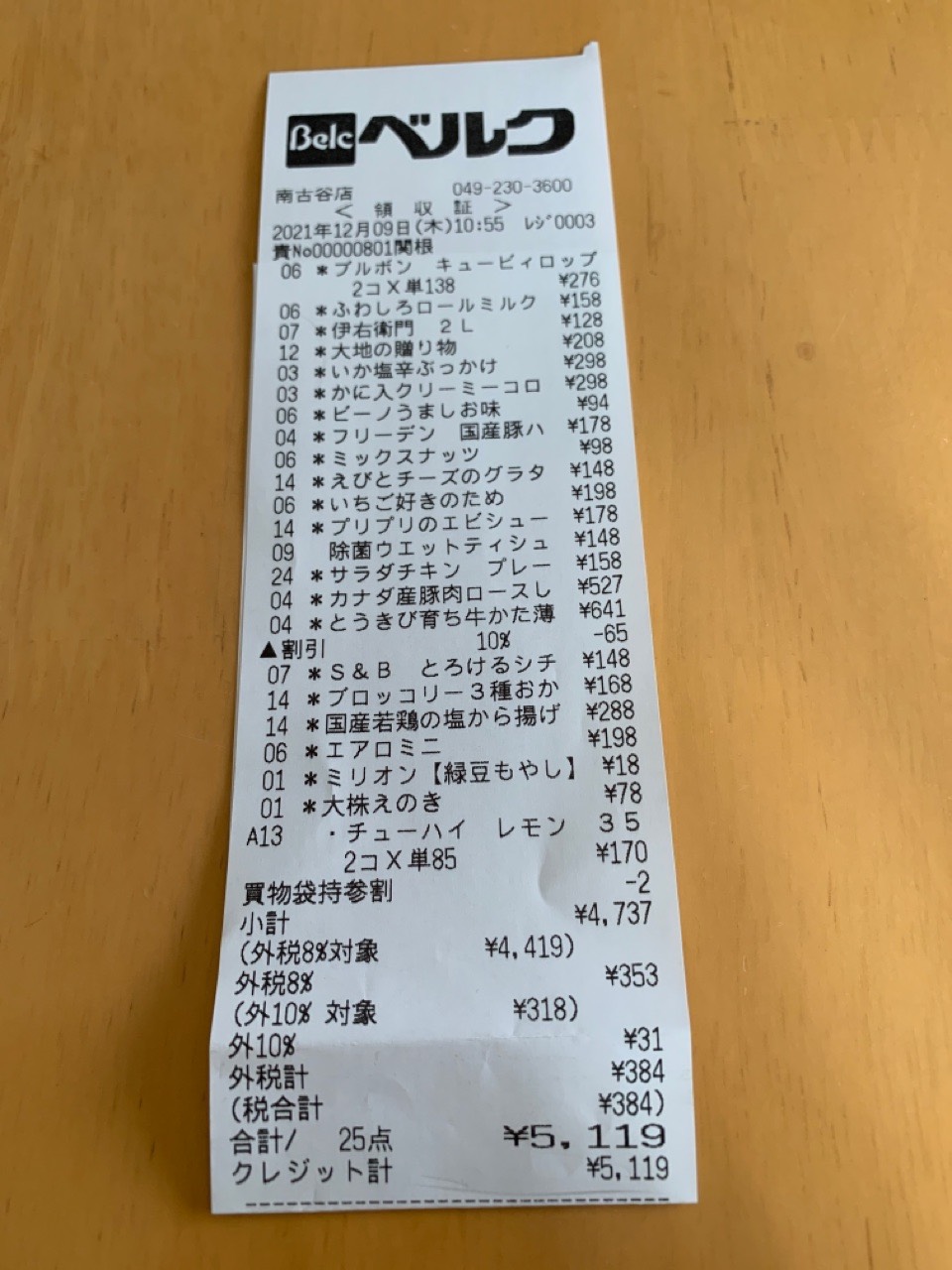}
  \includegraphics[width=2cm]{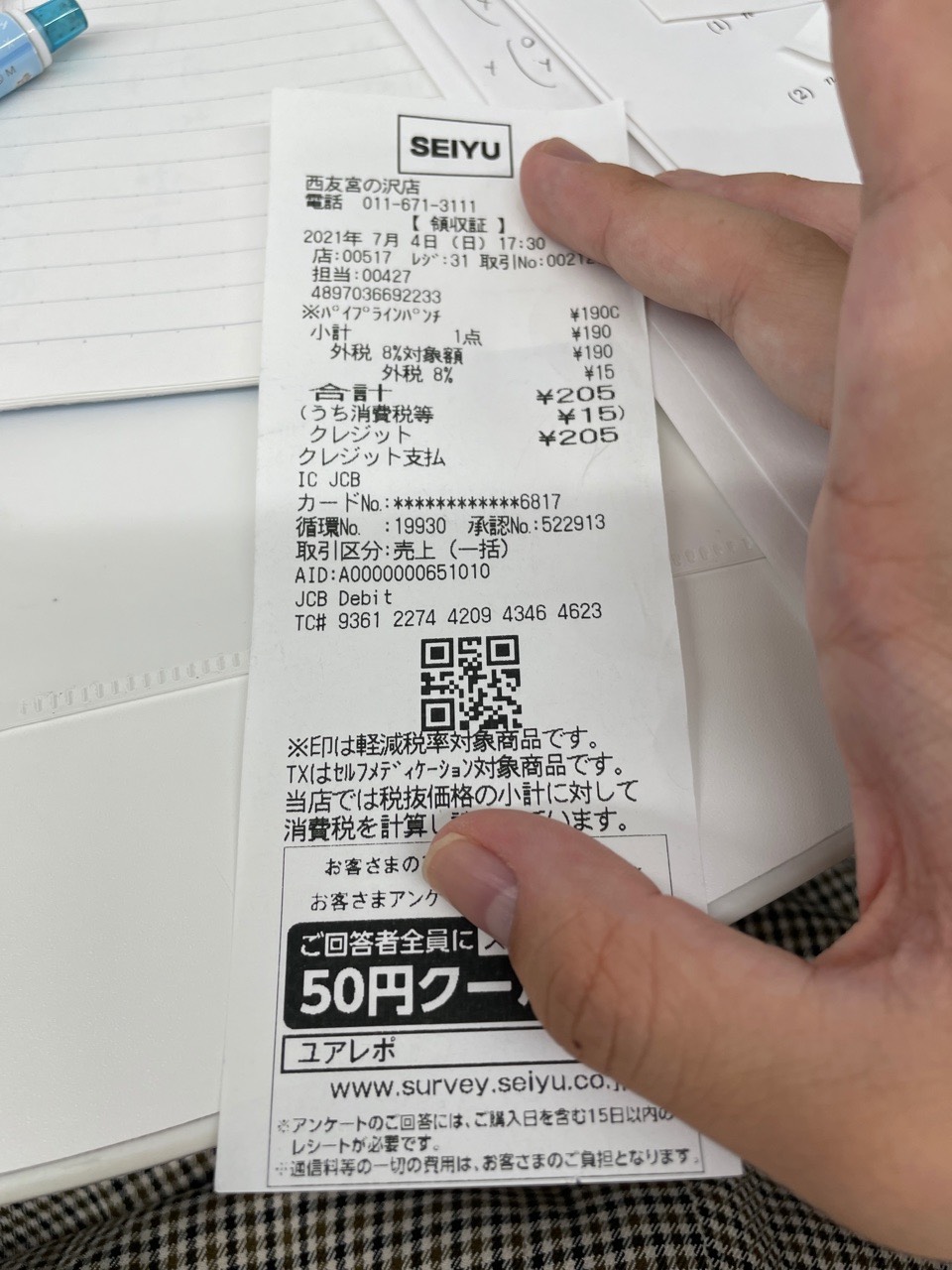}
  \caption{Eight real receipt images selected from the test set.}
  \Description{}
  \label{ReceiptExamples}
\end{figure}

For our experiments, we use receipt images collected by a third-party app. Each receipt image in our dataset was taken by a different user with 45\% coming from Android and 55\% from iOS. This helps ensure a wide variation in receipts, backgrounds and angles from which each image is taken. We select 1,225 receipt images in which all four corners are clearly visible, and manually annotate each image with a quadrilateral enclosing the four corners of the receipt. In Table~\ref{table1} we show the number of receipt images used for training, validation and testing. In Figure~\ref{ReceiptExamples} we show eight example receipt images taken from the test set.

\begin{table}
  \caption{Real data used for training, validation and testing.}
  \label{table1}
  \begin{tabular}{r|ccc}
    &Training&Validation&Testing\\
    \midrule
    Receipt Images &900 &100 &225\\
  \bottomrule
\end{tabular}
\end{table}

Each receipt image has one of the following pixel resolutions: 1080x1440, 1200x1600, 1440x1920, 1512x2016, 1536x2048, 2124x2832, 3024x4032, 960x1280.

\subsection{Synthetic data}\label{syntheticData}

In this section we describe in detail how we created the synthetic data:

\subsubsection{Receipt image selection}\label{Receipts}

\begin{figure}[h]
  \centering
  \includegraphics[height=4cm]{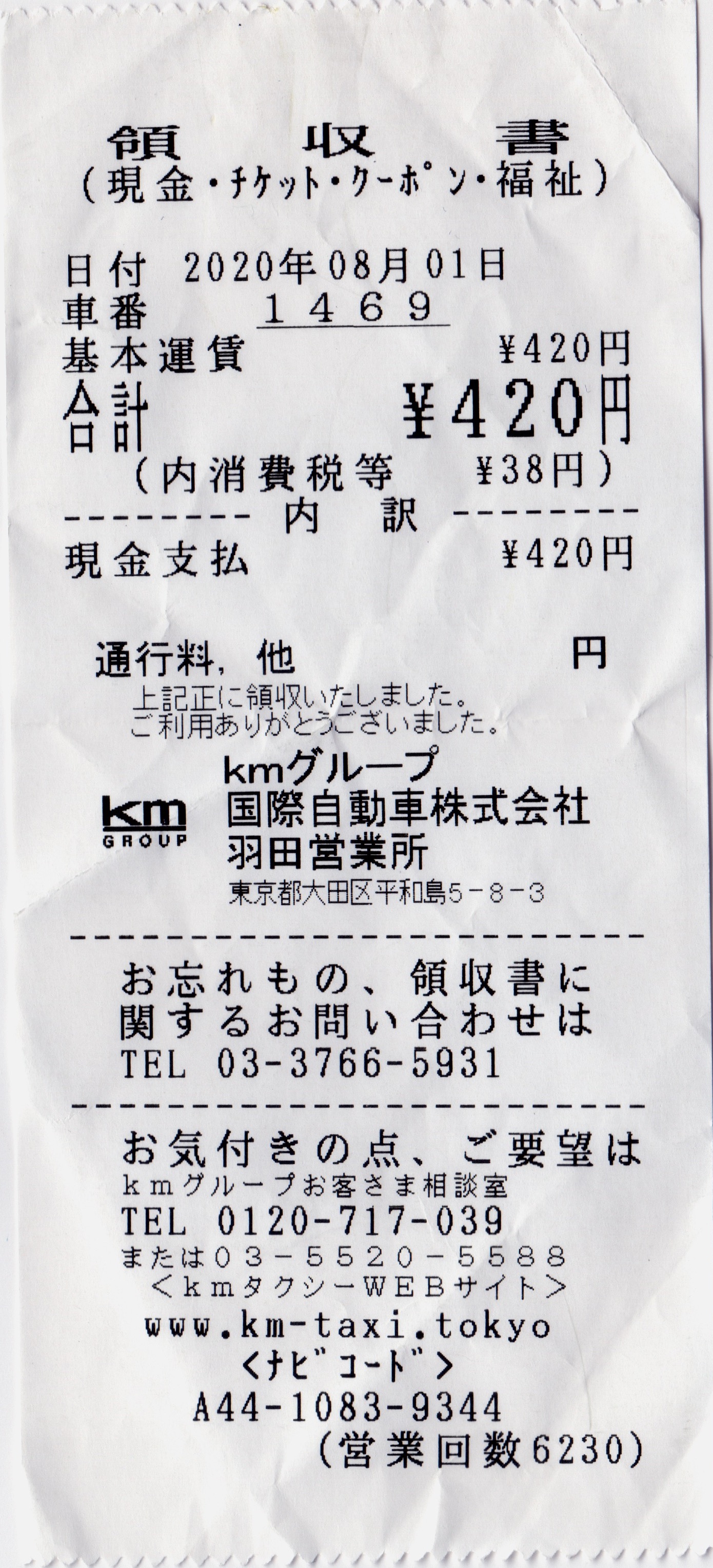}
  \includegraphics[height=4cm]{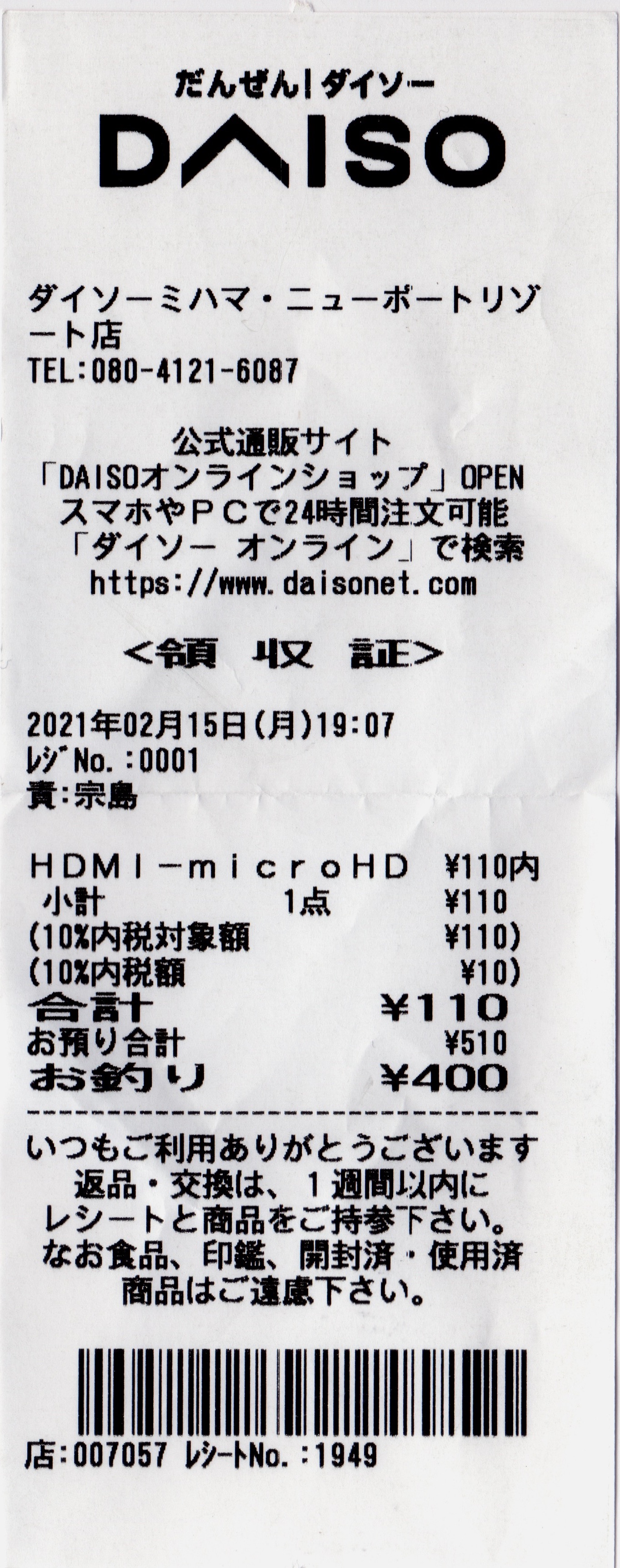}
  \includegraphics[height=4cm]{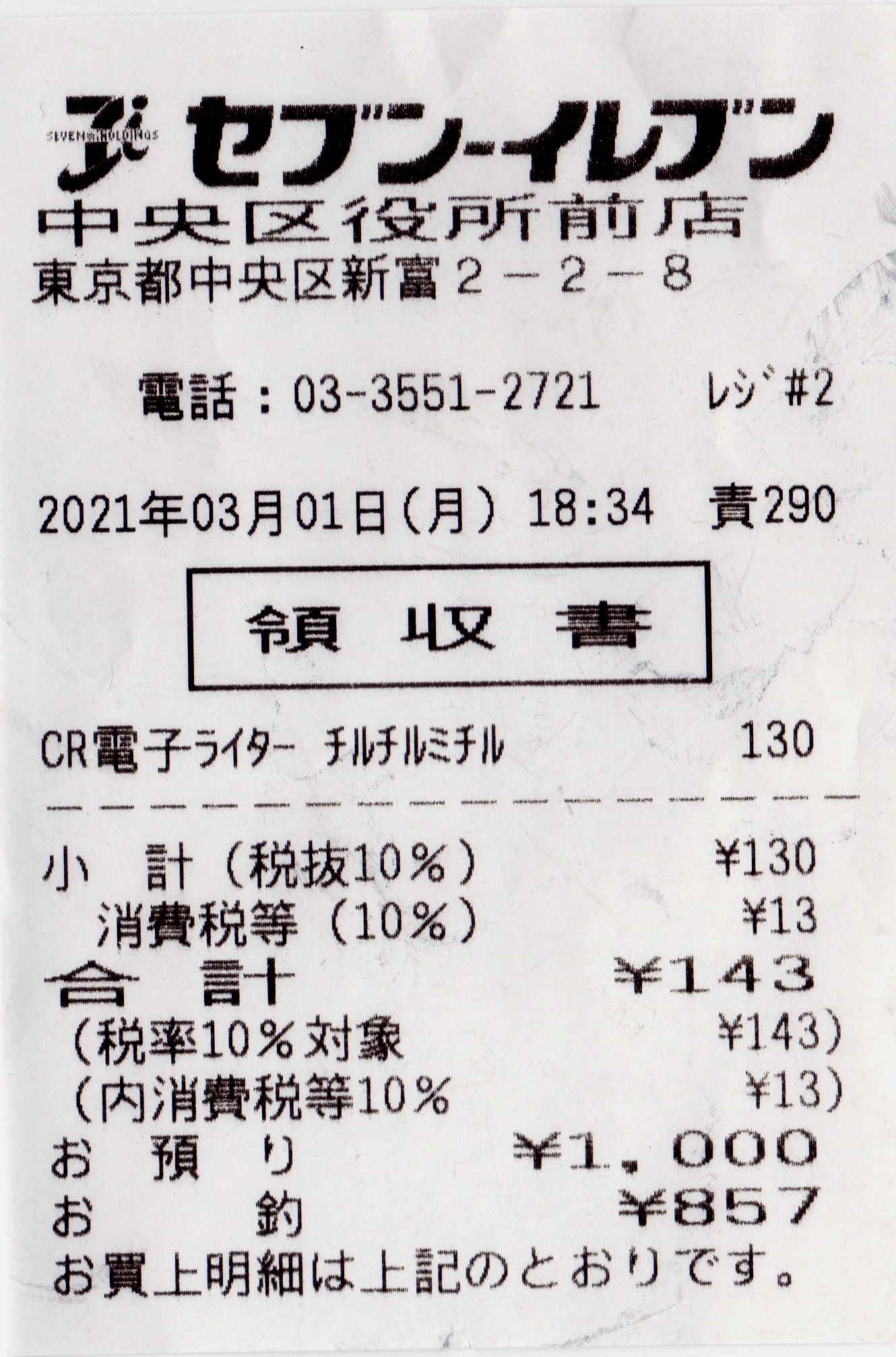}
  \caption{Three examples of the tightly cropped, portrait orientated receipt images used for generating our synthetic training data.}
  \Description{}
  \label{original_receipts_image}
\end{figure}

A dataset of approximately 10,000 scanned Japanese receipt images was obtained from Money Forward. Each receipt image was manually checked by the authors to ensure that the receipt in the image was in a vertical ``portrait'' orientation and that its edges and corners touched the image boundaries as closely as possible with a minimum amount of exposed background. Three example images from this data are shown in Figure~\ref{original_receipts_image}. This dataset is similar to, but much larger than, the Latin script-only receipt data available in the SROIE data\cite{Huang2019ICDAR2019CO}, although our crops are much tighter. We require close crops so that we can programatically superimpose these receipts on top of varied background images and know the location and identity of a receipt's corners so they can be automatically annotated with bounding boxes.

After discarding images which did not fit these criteria we ended up with 7,362 images, of which 7,204 are used exclusively for training and the remainder are not used in this work. In practice, the actual receipt contents, and indeed the fact that the text is a mixture of Japanese and Latin script, are unlikely to be important at the 300x300 resolution to which images are resized during training and inference. However, the rough appearance of blobs of text vs.~non-text are still likely to be important features, both for detecting a corner and also for identifying which corner it is. 

We will use the third image on the far right of Figure~\ref{original_receipts_image}, which shows the receipt for a purchase in a 7-11 convenience store in Japan, as the canonical example receipt in the following sub-sections, to illustrate the various transformations that are applied to all receipts to generate the synthetic images in our training data.

\subsubsection{Background image canvas selection}\label{Backgrounds}

While the possible variation of receipts is large, their main features may be considered essentially finite, especially at the 300x300 pixel resolution used by our SSD object detection model. In contrast, the variation in potential backgrounds is infinite: as is usual with text-in-the-wild we have no idea where, or how, a user will place their receipts when taking a picture of them. We choose a wide range of possible backgrounds as canvases for our training images, and hope that these will be representative. We expect that increasing the number and variation of backgrounds, especially those with relatively low contrast compared to the receipts themselves, and containing features that might be confused with receipts, are likely to produce better models. 

\begin{figure}[h]
  \centering
  \includegraphics[width=\linewidth]{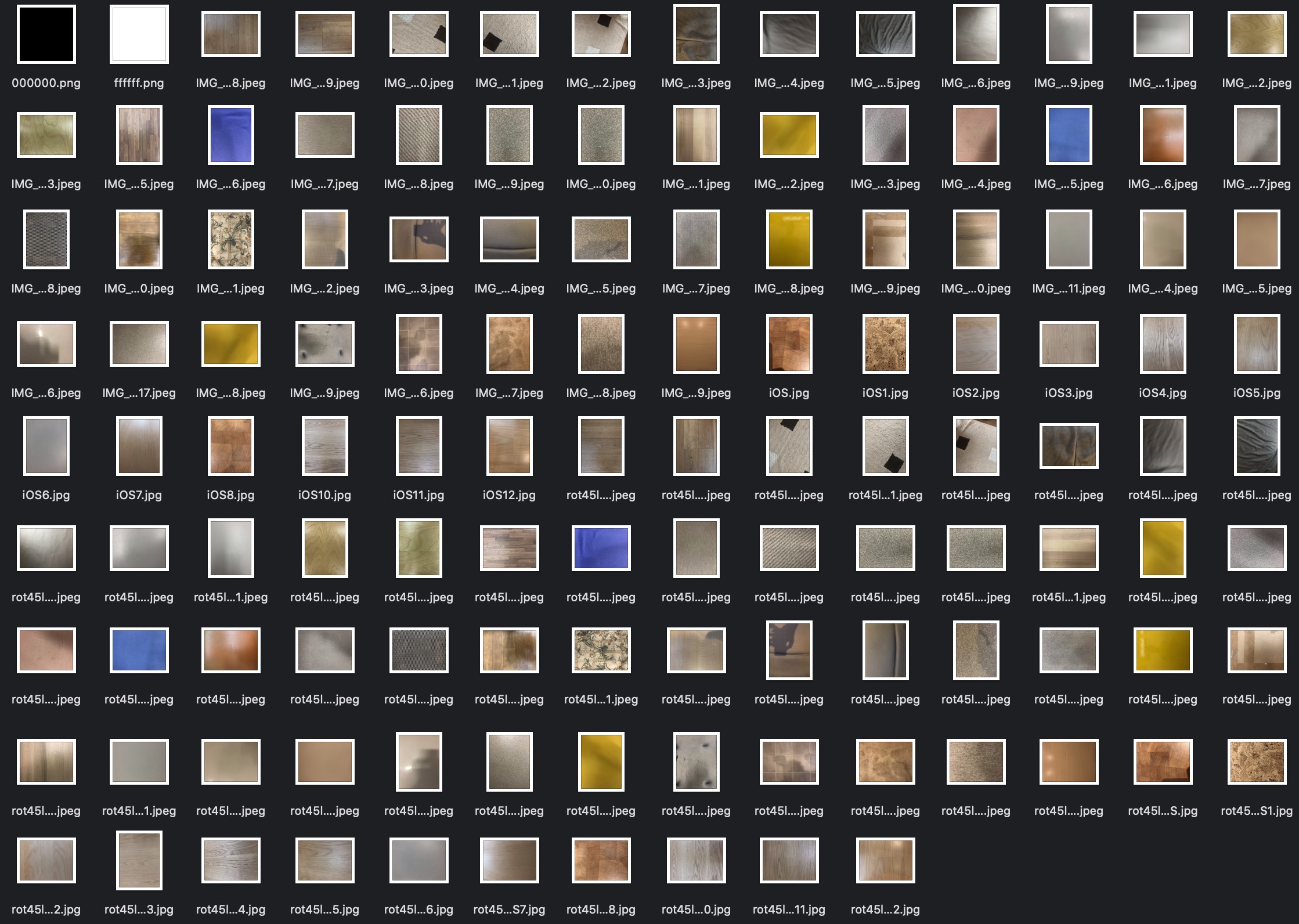}
  \caption{The 122 background canvas images on which synthetic receipt images were then superimposed.}
  \Description{.}
  \label{background_images}
\end{figure}

An icon list layout of the 122 background canvas images used for generating the synthetic training data is shown in Figure~\ref{background_images}.
These background images were generated using either the  OpenCV\cite{opencv_library} computer vision library to create constant-pixel-value images (e.g. ``pure'' black, white and gray images), or an iPhone to take photos of large objects with homogeneous color and texture, such as household tables and carpets. Each background image was also rotated by 90 degrees and added to the set of backgrounds.

Background canvas images are then resized to 1080x1920 pixels which is a similar aspect ratio to the typical receipt when being captured in a portrait orientation.

\subsubsection{Generating synthetic images}

Synthetic images are created by superimposing a single target receipt image, as described in Section~\ref{Receipts}, at the center of one of the background canvas images, described in Section~\ref{Backgrounds}. Optionally, up to two interfering receipt images are also added to the synthetic image.

Due to the way in which augmentation is performed in subsequent steps, when an image is rotated, shifted and otherwise transformed, it is highly likely that the original bounding boxes will no longer match the region of the image to which they have been transformed. For our usage case, we only need to know the location of bounding boxes on the corners of each receipt. By keeping these bounding boxes as small as possible we can ensure that image transformations cause the transformed bounding boxes to much more accurately match the resulting image corners after transformation. In the limit, a bounding box that is 1 pixel in size will move to the correct location after augmentation. For any other size of bounding box the center of the bounding box after augmentation is not guaranteed to be the center of the bounding box after augmentation. We use a square bounding box of side length 10 pixels centered on each corner to minimise this potential discrepancy.

Before superimposing the receipt image on the background canvas image we resize it so the relative size of the receipt in the image frame is similar to what we expect in reality. The size of the receipt might be affected by subsequent augmentations but we need to be careful that the size is such that receipts in an augmented image do not go over the background canvas boundary. 

It is particularly important to note that the model which is trained on these images might pick up on unintended trends in our training data. For example, if all the receipts in an image are very small the final model might only detect receipt corners when it sees a similarly small receipt. This would then mean that receipts will only be detected when a user holds their smartphone far from the target receipt. This is almost certainly not what we want, since, even if the receipt is correctly detected, once it is extracted and rectified there may be too few pixels to recognize what is actually printed on the receipt. We therefore resize the receipt image such that there is a margin above and below the receipt of 30\% of the height of the background image.

Lastly, we specify a maximum of two extra receipts that will be randomly added to the background canvas image. During our initial investigation of this approach, we only created training images containing a single target receipt. However, it was found that this early model had trouble identifying the target receipt when other receipts were also visible, which is quite likely to occur in practice. Therefore, we add a random number of ``interfering'' receipts (between zero and two) in random positions, with random orientations relative to the target receipt, both surrounding and under the target receipt. Note that we do not add receipts on top of the target receipt. As expected, using this modified data for training resulted in receipt detection that was much more robust to interfering receipts than the initial model. 

\begin{figure}[h]
  \centering
  \includegraphics[height=5cm]{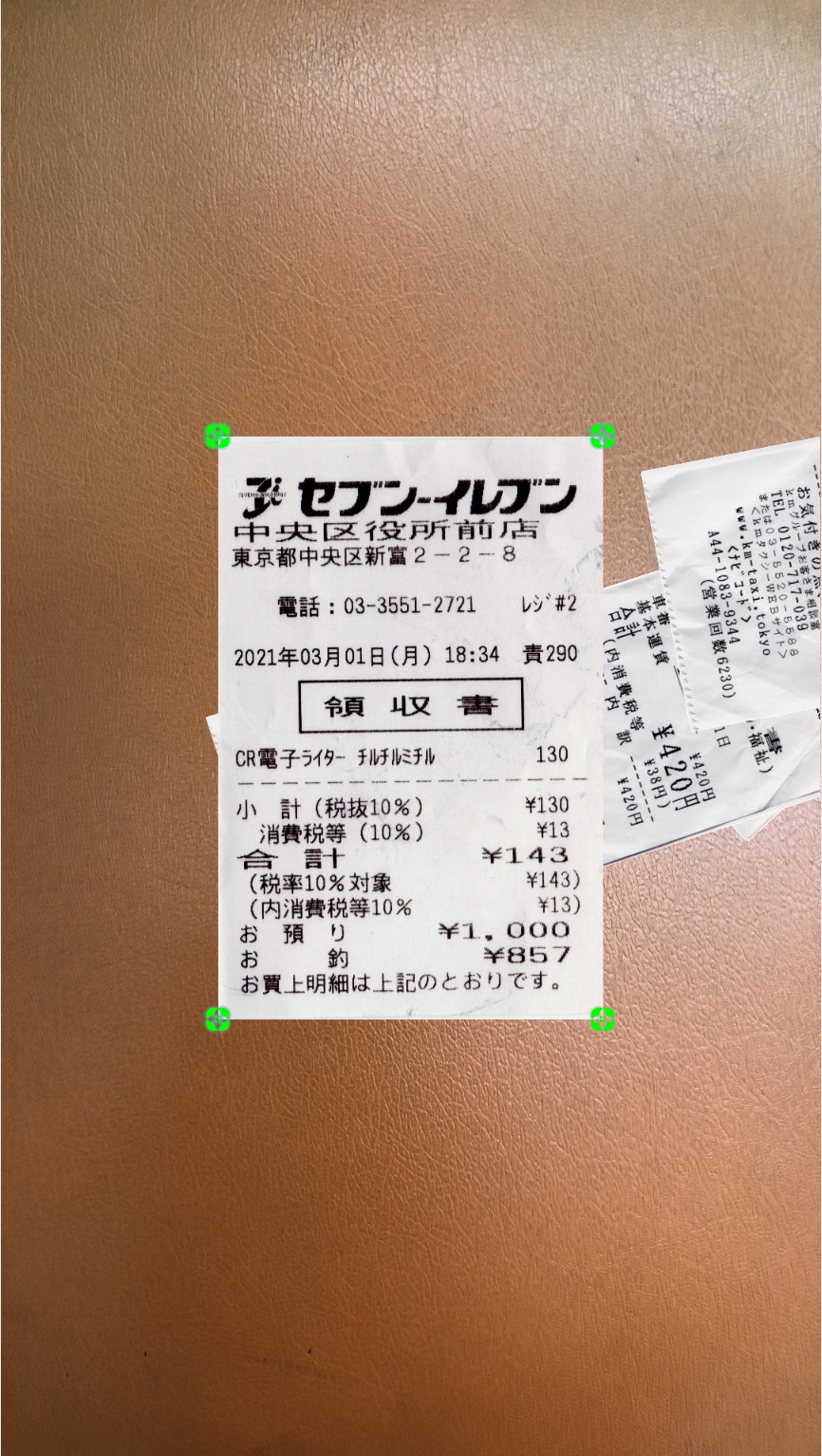}
  \caption{An automatically generated synthetic image showing a random selection of three sample receipts on a randomly selected background image, where the two ``interfering'' receipts are positioned underneath, and oriented randomly with respect to, the centrally positioned and vertically oriented target receipt.}
  \Description{.}
  \label{rectlinear_synthetic_receipts}
\end{figure}

In Figure~\ref{rectlinear_synthetic_receipts} we show how the original receipt in its original orientation shown in the far right of  Figure~\ref{original_receipts_image} has been placed on top of an orange background image, and two additional ``interfering'' receipts with random rotations have been added underneath it. Figure~\ref{rectlinear_synthetic_receipts} also shows in bright green the location of the 10-pixel square bounding boxes on each corner of the target receipt. Note that there are no bounding boxes annotated on any of the interfering receipts. Each of the four corners is assigned a unique object label as follows: \texttt{tl}, \texttt{tr}, \texttt{br} and \texttt{bl}, corresponding to the \textit{top-left}, \textit{top-right}, \textit{bottom-right} and \textit{bottom-left} corners, respectively.

\subsubsection{Augmenting receipts}

While the ``bird's-eye-view'' shown in Figure~\ref{rectlinear_synthetic_receipts} might sometimes be encountered in reality it is nonetheless very constrained and goes against our original objective which is to make receipt detection and extraction as eﬀicient and robust as possible, irrespective of elevation, roll, inclination or rotation of the camera with respect to the receipt being imaged. Consequently, we need to simulate the situation where the camera views the target receipt from a large variety of different perspectives. Once again we perform this operation automatically by applying various random rotations, shifts and projective transforms to the images prepared in the previous step. Specifically, we use the \textit{imgaug} image augmentation tool\cite{imgaug} available on GitHub\footnote{\texttt{https://github.com/aleju/imgaug} (version 0.3.0)} and apply the following image transformations: 

(1) scale the image randomly by a factor of between 0.5 and 1.0 in both the $x$ and $y$ directions; (2) translate the image both left-right and up-down randomly by between -30\% and 30\% of the original image size; (3) rotate the image about its center randomly by between -70 and 70 degrees; and (4) apply a projective transformation of between 0 and 0.15---this value is ``roughly a measure of how far the projective transformation’s corner points may be distanced from the image’s corner points''\cite{imgaug} while keeping the size of the image the same. We ensure that the transformed image stays within the original image boundaries and any previously non-visible pixels (i.e.~those that were originally off the edge of the original image), that come into view after transformation, are given values which match the closest visible original pixel value (color) to that pixel prior to transformation. In general, this means that we assign new pixels the same color as whatever the color at the edge of an image was. 

\begin{figure}[h]
  \centering
  \includegraphics[height=5cm]{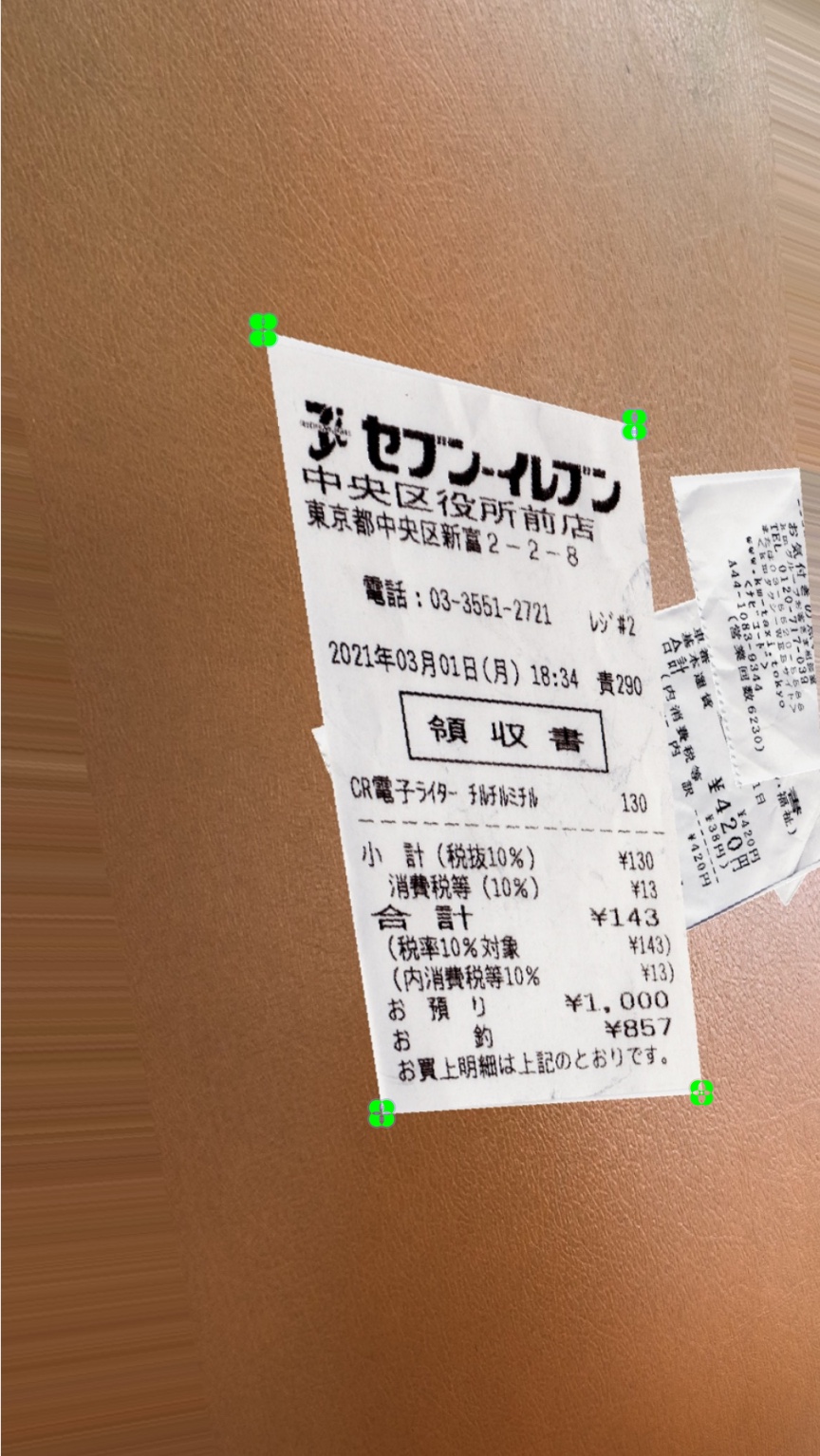}
  \caption{The same image from Figure~\ref{rectlinear_synthetic_receipts} showing one possible projective transformation. In particular, note how the corners appear after the change of perspective.}
  \Description{.}
  \label{synthetic_after_augmentation}
\end{figure}

One example of applying the above transformation routine to the image from Figure~\ref{rectlinear_synthetic_receipts} is shown in Figure~\ref{synthetic_after_augmentation}.
Comparing these two images we can easily imagine that the same physical receipts are present but that the camera has been moved to a different physical position, relative to the receipts. This, of course, is exactly what we are trying to simulate.

\subsubsection{Bounding box expansion}\label{bbox_expansion_explanation}

At this point, for each original image, we also have two augmented versions of it. The annotations for all of these images are still the small bounding boxes on the corners of each target receipt, as shown in Figure~\ref{synthetic_after_augmentation}. If we were to leave the bounding boxes as small as this it is unlikely that the model would learn sufficiently discriminating features for the model to be able to determine which corner is which. This will be even more problematic once the images have been resized to 300x300 pixels. Moreover, the model is more likely to be confused by other small objects in the image that happen to look similar to these small receipt corner regions, especially at such low resolutions.

An important point to remember here is that ultimately we want to know as precisely as possible where each corner of the receipt is. As we mentioned earlier, a receipt corner is defined to be the centre of the bounding box, which is itself defined to be square. We know that during inference the accuracy of the location of bounding boxes will be dependent on many things but in general we can assume that the larger the bounding box is, the closer it will be to its correct location. Thus to minimize this location error we want all bounding boxes to be as large as possible. We therefore expand each bounding box up to its maximum size (i.e.~up to the closest edge of the image) while preserving its square proportion by keeping all sides the same length. 

We speculate that this has the additional benefit that we are likely to capture more informative features such as strong diagonals radiating out from the centre of the bounding box along the edge of each receipt, as well as capturing text and non-text features that might distinguish a receipt from its background.

\begin{figure}[ht]
  \centering
  \includegraphics[height=5cm]{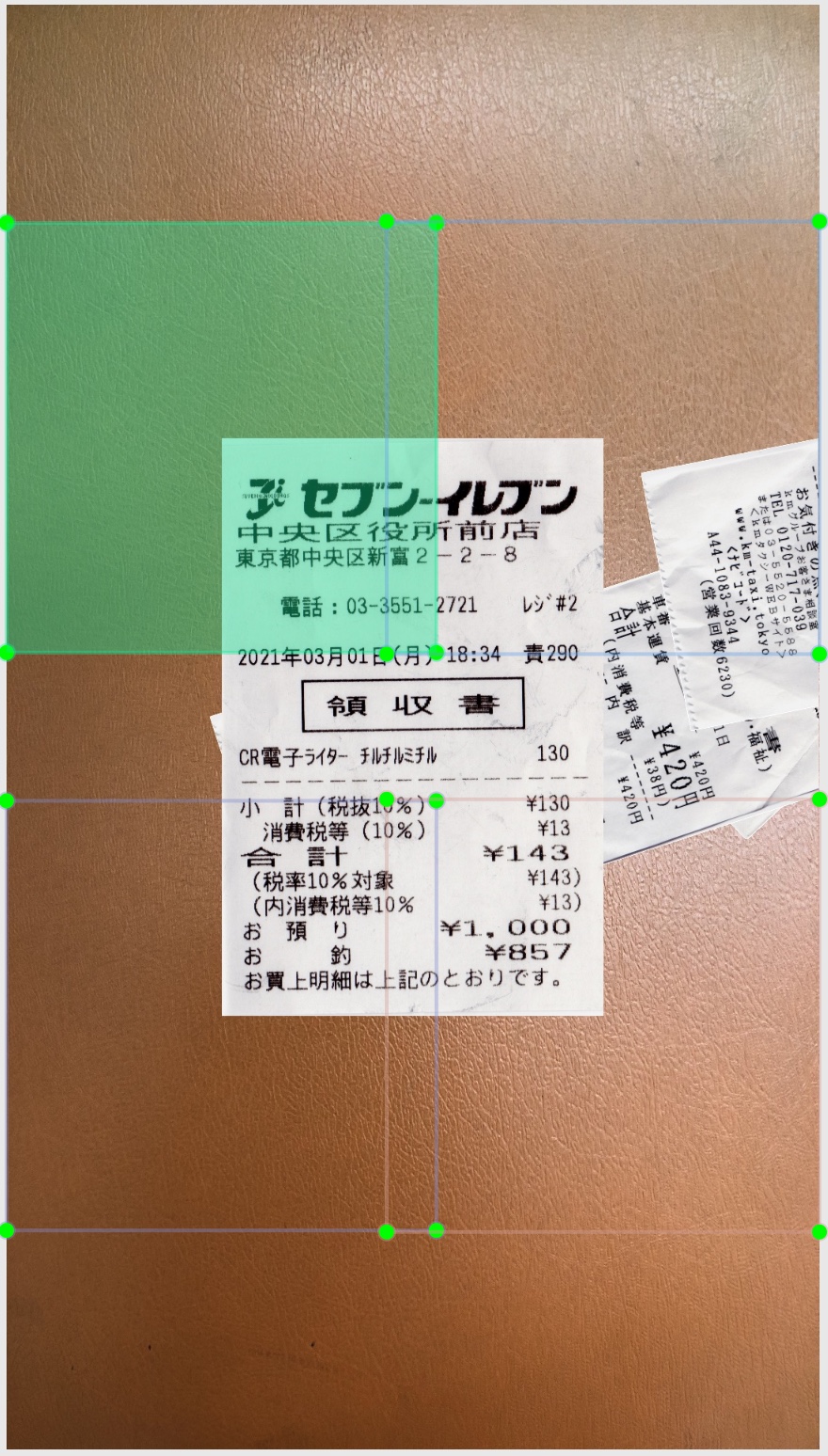}
  \includegraphics[height=5cm]{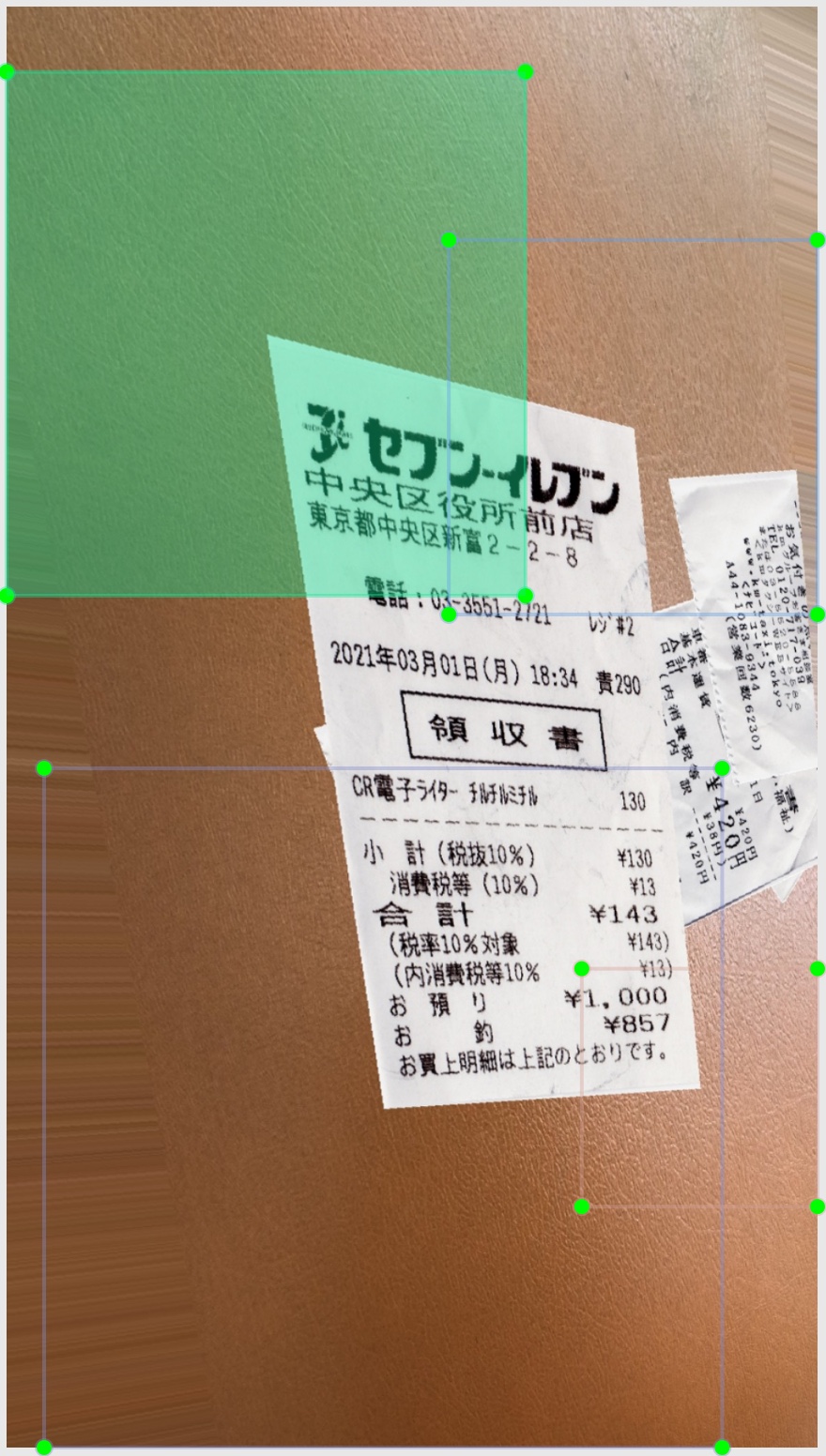}
  \caption{Image and annotations displayed using the LabelImg annotation tool, highlighting the expanded corner bounding boxes (bright green) with the \texttt{tl} bounding-box filled (pale green), showing its maximal expansion to the closest image boundary while maintaining a square shape.}
  \Description{}
  \label{image_showing_expanded_bounding_boxes}
\end{figure}

For the two images shown in Figures~\ref{rectlinear_synthetic_receipts} and~\ref{synthetic_after_augmentation}, in Figure~\ref{image_showing_expanded_bounding_boxes} we display the bounding box annotations as they appear in the  LabelImg\footnote{\texttt{https://github.com/tzutalin/labelImg}} annotation tool, using light green to denote the four expanded bounding boxes on the four corners of the target receipt, with only the bounding box for the \texttt{tl} receipt corner shown filled in.

At the end of this generation procedure we have 21,612 combined receipt and background synthetic images to train with. Note that we do not use any synthetic data for testing or validation.

\section{Experiments}\label{Experiments}

\subsection{Baseline approach}

State-of-the-art baselines for this task are difficult to obtain since they are typically proprietary. We therefore choose the implementation using edge detection that is available from PyImageSearch\footnote{\texttt{https://www.pyimagesearch.com/2014/09/01/\\build-kick-ass-mobile-document-scanner-just-5-minutes/}} since it is specifically aimed at detecting receipts and uses only traditional computer vision techniques available in the OpenCV library. While we appreciate that this implementation does not necessarily represent the state-of-the-art, we believe it gives a reasonable baseline against which to compare our proposed method. Perhaps more importantly it provides a readily available, easy-to-use, implementation that others may also compare with.

There are several parameters that can be tuned with this method, as described in Section~\ref{Baseline_Approach}. We tune these parameters using an exhaustive grid search on the validation data, with all combinations of the following parameters between the given limits and increments investigated: Gaussian kernel size from 1 to 21, in steps of 2; Canny thresholds from 0 to 500 in steps of 50; Epsilon of the Douglas-Peucker polygon approximation algorithm from 0.01 to 0.1 in steps of 0.01.

We select the set of parameters which gives the best accuracy on the validation data which results in: Gaussian kernel=5; Canny threshold1=50; Canny threshold2=0; Epsilon=0.06.

\subsection{Proposed approach}

We train two SSD MobileNet object detection models using two different training datasets: (1) only the real training data described in Sub-section~\ref{RealData}; and (2) both the real training data, and the synthetic data described in Sub-section~\ref{syntheticData}. Each model has its weights initialized randomly and is trained using the Tensorflow Machine Learning Toolkit~\cite{tensorflow2015-whitepaper} v2.7.0 and the Object Detection API. 

We use the default settings in the configuration file for the SSD MobileNet model trained on the COCO 2017 Dataset named \texttt{ssd\_mobilenet\_v2\_320x320\_coco17\_tpu-8} which is available from the Tensorflow Model Zoo\footnote{\texttt{http://download.tensorflow.org/models/object\_detection/tf2/20200711/\\ssd\_mobilenet\_v2\_320x320\_coco17\_tpu-8.tar.gz}} such as learning rate (0.8 with a cosine decay for the first 50,000 iterations) and batch size of 24. Note that we are not performing transfer learning; only the default configuration file is used. We train each model until the mean average precision (mAP) at 0.5 intersection-over-union (IoU) no longer increases on the validation data.

\section{Results}\label{Results}

As explained in Section~\ref{bbox_expansion_explanation} our models are trained using bounding box annotations where each box is centred on a corner and then expanded to the nearest image boundary while maintaining a square shape. We expect that the model will learn this definition given enough data although there is no guarantee. We therefore avoid the use of IoU as our evaluation metric, since the size of each box is affected by its position, in particular its distance from the closest image boundary. Instead we adopt a stricter definition: the euclidean distance in pixels of the predicted corner from the manually annotated reference corner. We then set a threshold distance for a predicted corner to be accepted as correctly detected and evaluate performance for a range of thresholds from 10 to 50 pixels. Since each image has a resolution of between 960x1280 and 3024x4032 these thresholds correspond to a tolerance of between 0.25\% and 1\% at threshold 10, and 1.25\% and 5\% at threshold 50.

It should be noted that there is in any case a tolerance on the accuracy with which a human annotator can accurately identify the corner of a receipt. So we believe this metric is suitable for our purpose. For rectification we typically require a tighter threshold since deviations in identifying the receipt corners can result in highly distorted rectified images that are unusable. However, depending on the downstream processing task, larger thresholds (i.e. greater distances of the predicted corner from the correct corner) might also be acceptable. 

In Table~\ref{table2} we show the results using five different thresholds of our proposed corner detection method, trained on real data only (real only), and both real and synthetic data (real+syn), against the baseline method, which is not trainable but is optimised on real data. We evaluate both the total percentage of corners in the dataset that are detected correctly, as well as the percentage of receipts for which all four corners are correctly detected. The latter metric is more relevant since a method that detects fewer than four corners is not really useful.

\begin{table}
  \caption{Percentage of correctly identified corners and correctly identified receipts (all four corners detected correctly) in our test set using each method at different thresholds.}
  \label{table2}
  \begin{tabular}{l|r|ccccc}
        \multicolumn{2}{c|}{}&\multicolumn{5}{|c}{\bfseries Euclidean pixel distance threshold}\\
        \multicolumn{2}{c|}{}&10 & 20 & 30 & 40 & 50\\
    \midrule
    Baseline &Corners     &34.6\% &38.8\% &39.4\% &41.6\% &42.1\%\\
             &Receipts    &23.1\% &33.8\% &34.2\% &36.0\% &36.9\%\\
    \midrule
    Proposed &Corners     &77.3\% &90.0\% &92.3\% &93.2\% &93.9\%\\
    (real only)&Receipts  &44.0\% &72.4\% &76.9\% &79.1\% &80.9\%\\
    \midrule
    Proposed &Corners     &\textbf{80.1}\% &\textbf{91.2}\% &\textbf{94.2}\% &\textbf{94.7}\% &\textbf{95.1}\%\\
    (real+syn)&Receipts   &\textbf{49.8}\% &\textbf{74.2}\% &\textbf{82.7}\% &\textbf{84.0}\% &\textbf{85.3}\%\\

  \bottomrule
\end{tabular}
\end{table}

\section{Discussion}\label{Discussion}

In Table~\ref{table2} we see that our proposed method shows a very clear improvement over the baseline with the total percentage of corners detected by the proposed method of between 80.1\% and 95.1\% depending on the acceptance threshold. This compares favorably to the baseline method which shows an accuracy of between 34.6\% and 42.1\%. The stricter metric, which evaluates whether all four corners are correctly detected, gives an accuracy of between 49.8\% and 85.3\%, compared to the baseline method which showed an accuracy between 23.1\% and 36.9\%. 

These results also show that using both the real and synthetic training data improves the accuracy of the model trained only on the real data, by between 5.4\% and 13.1\% relative.

It is interesting to note that the default values used in the PyImageSearch implementation of Gaussian kernel=5; Canny threshold1=75; Canny threshold2=200; Epsilon=0.02 gave a receipt detection performance of only 32.4\% when scored with threshold=50. This shows that the parameter optimisation on the held-out validation dataset improved performance on our test set over the default implementation by almost 14\% relative to 36.9\%. Nonetheless, even this optimised implementation of the baseline method was unable to detect four corners in 21\% of cases.

The data we used for testing is characterized by a wide variety of receipt types and shapes (folds and creases) in a variety of backgrounds, with a variety of interfering objects, across the two main smartphone OSes. Our synthetic data was prepared with the objective of identifying the corners of a single central target receipt against a potentially low-contrast (i.e.~typically the same white-ish color) background, in which possibly multiple ``interfering'' receipts are also present. Such low-contrast receipts made up 15\% of the images in our test set. 

\begin{figure}[h]
  \centering
  \includegraphics[width=2cm]{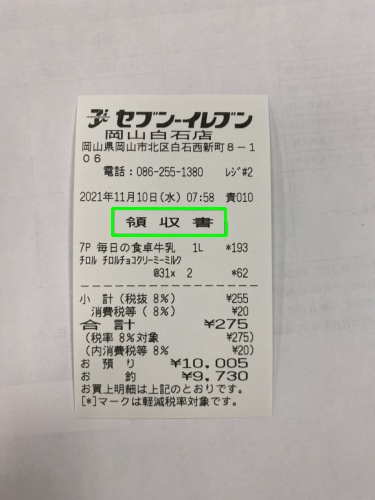}
  \includegraphics[width=2cm]{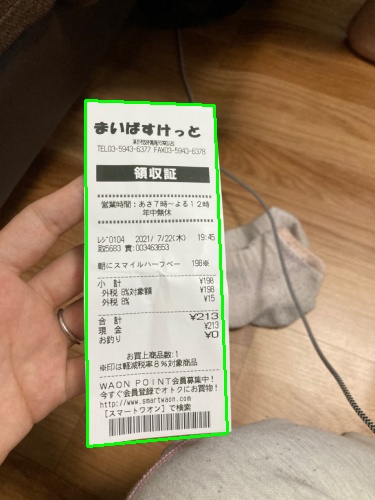}
  \includegraphics[width=2cm]{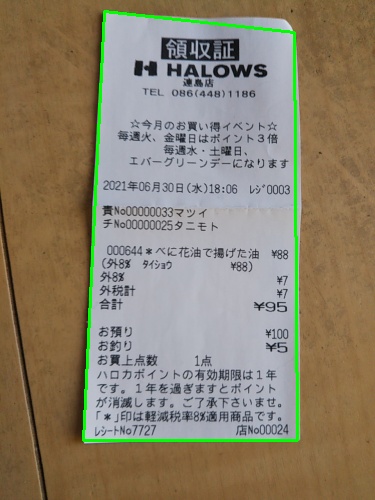}
  \includegraphics[width=2cm]{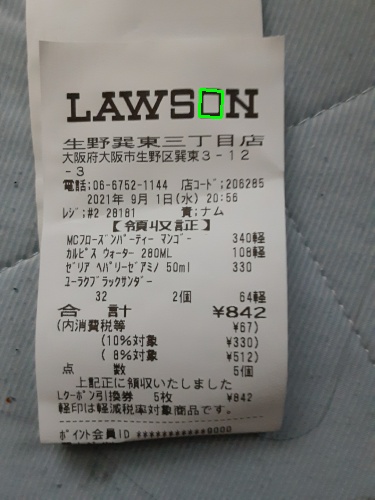}
  \includegraphics[width=2cm]{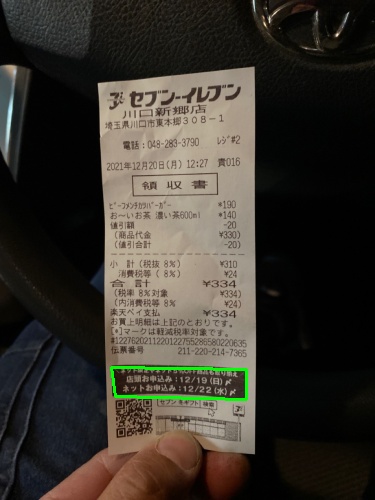}
  \includegraphics[width=2cm]{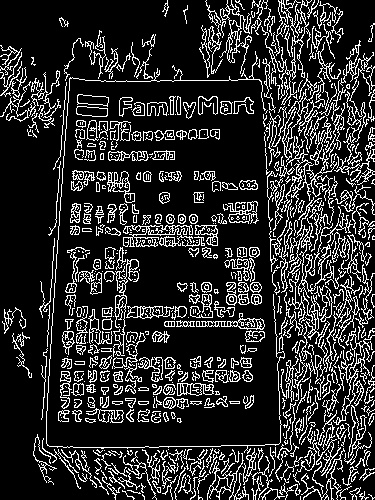}
  \includegraphics[width=2cm]{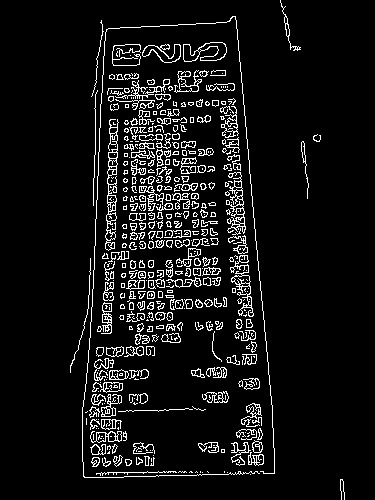}
  \includegraphics[width=2cm]{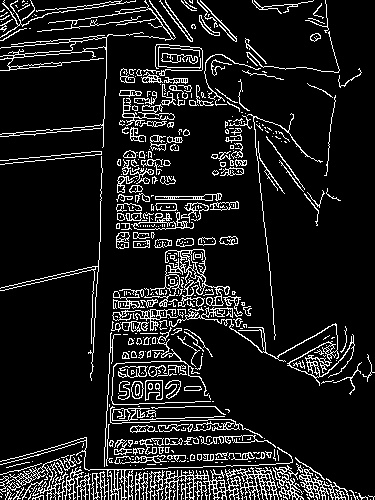}
  \caption{The eight example receipt images shown in Figure~\ref{ReceiptExamples} after automatic detection using the baseline method tuned on real validation data and showing the detected receipt regions. For the three images at bottom-right no region was detected so we show the edge detection results for those images instead.}
  \Description{}
  \label{ReceiptExamplesAfterDetectionWithBaseline}
\end{figure}

In Figure~\ref{ReceiptExamplesAfterDetectionWithBaseline} we show the results of using the tuned baseline method to perform corner detection on the example receipts shown in Figure~\ref{ReceiptExamples}. For the three receipts at the bottom-right no quadrilateral was found using the baseline approach, so instead we display the edge detection results. To highlight how fragile the traditional edge detection approach can be, compare these images with the center image in Figure~\ref{edge_detection_example} which used the default parameters and successfully located the correct quadrilateral. 

\begin{figure}[h]
  \centering
  \includegraphics[width=2cm]{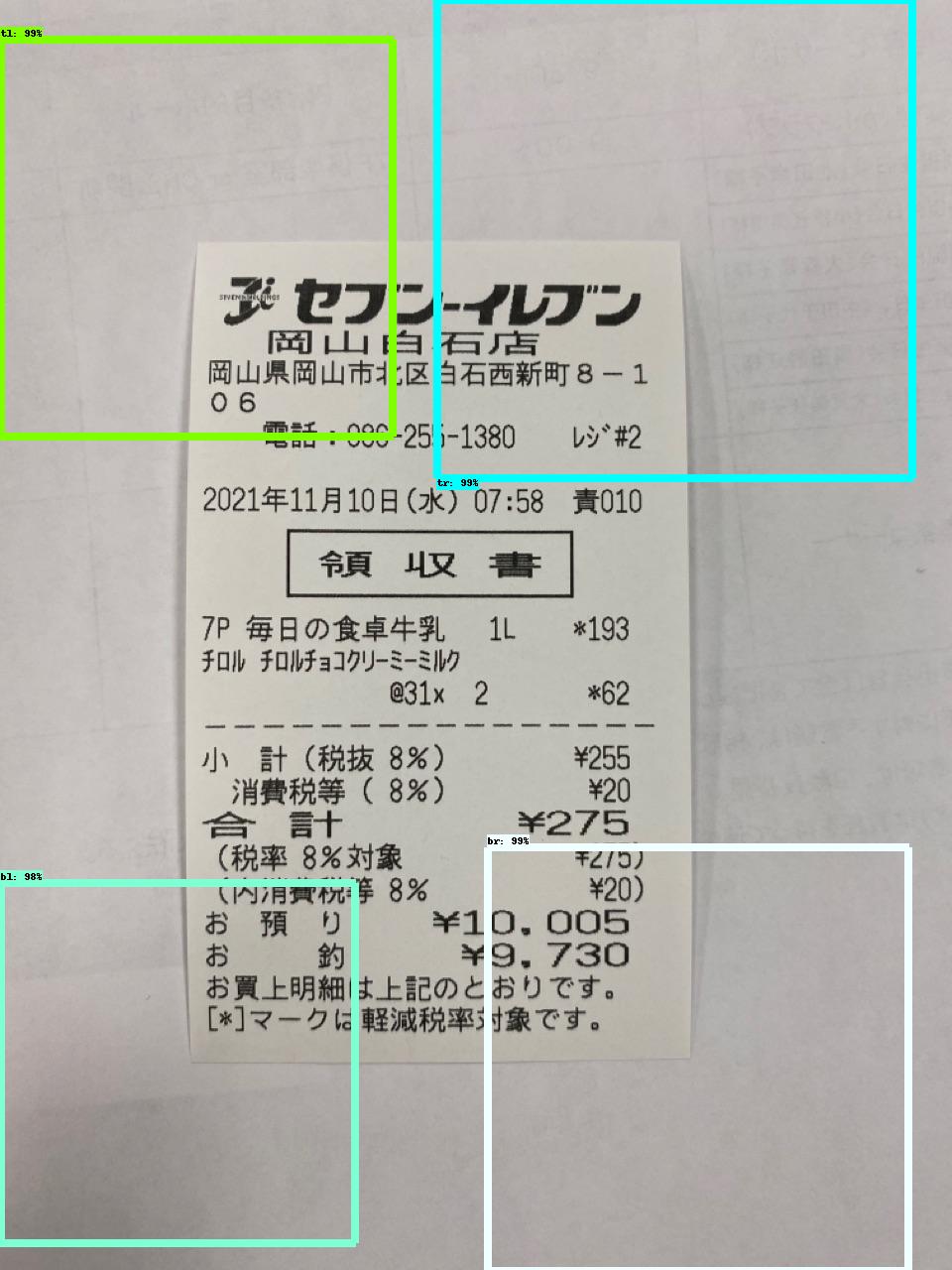}
  \includegraphics[width=2cm]{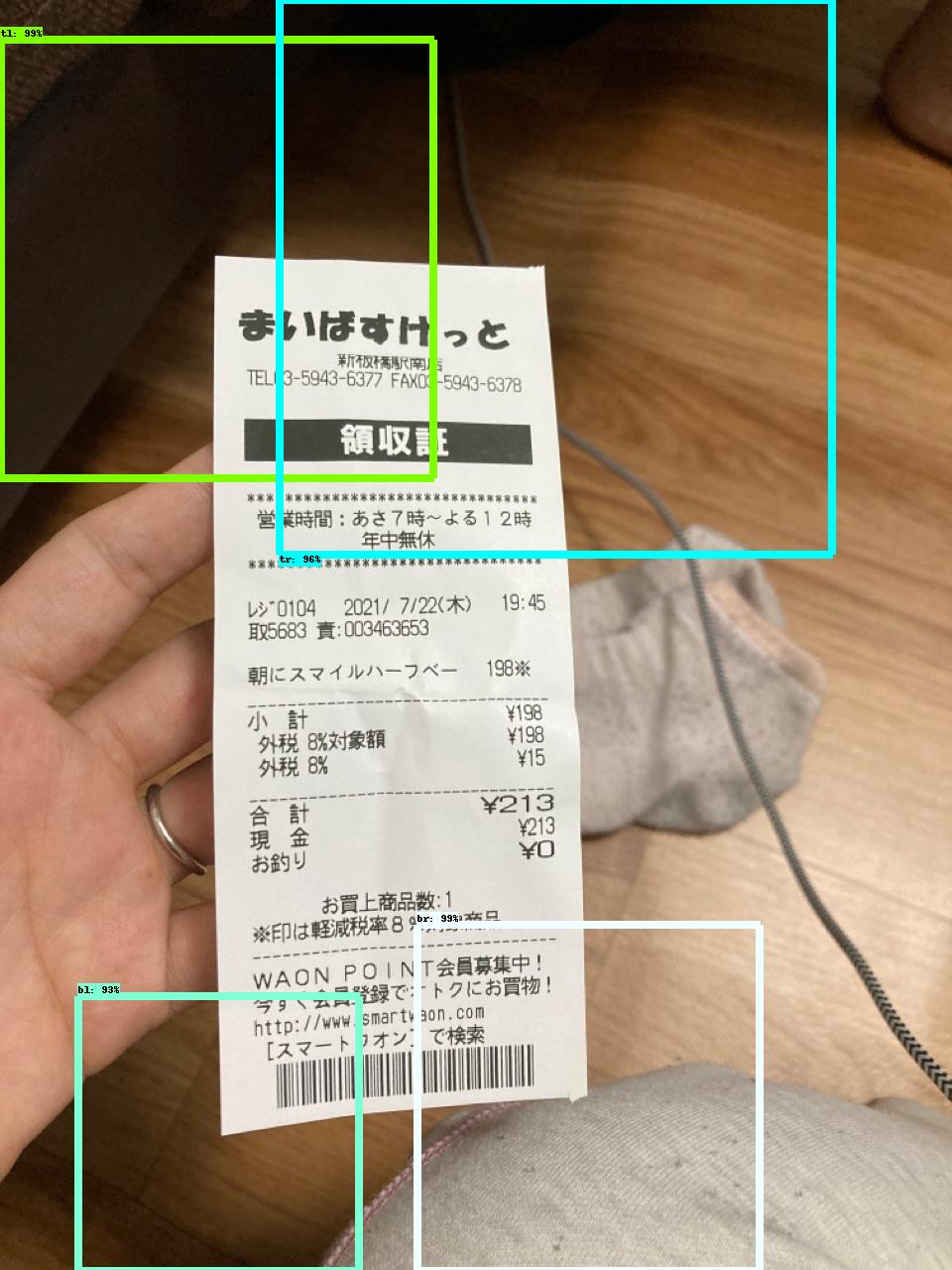}
  \includegraphics[width=2cm]{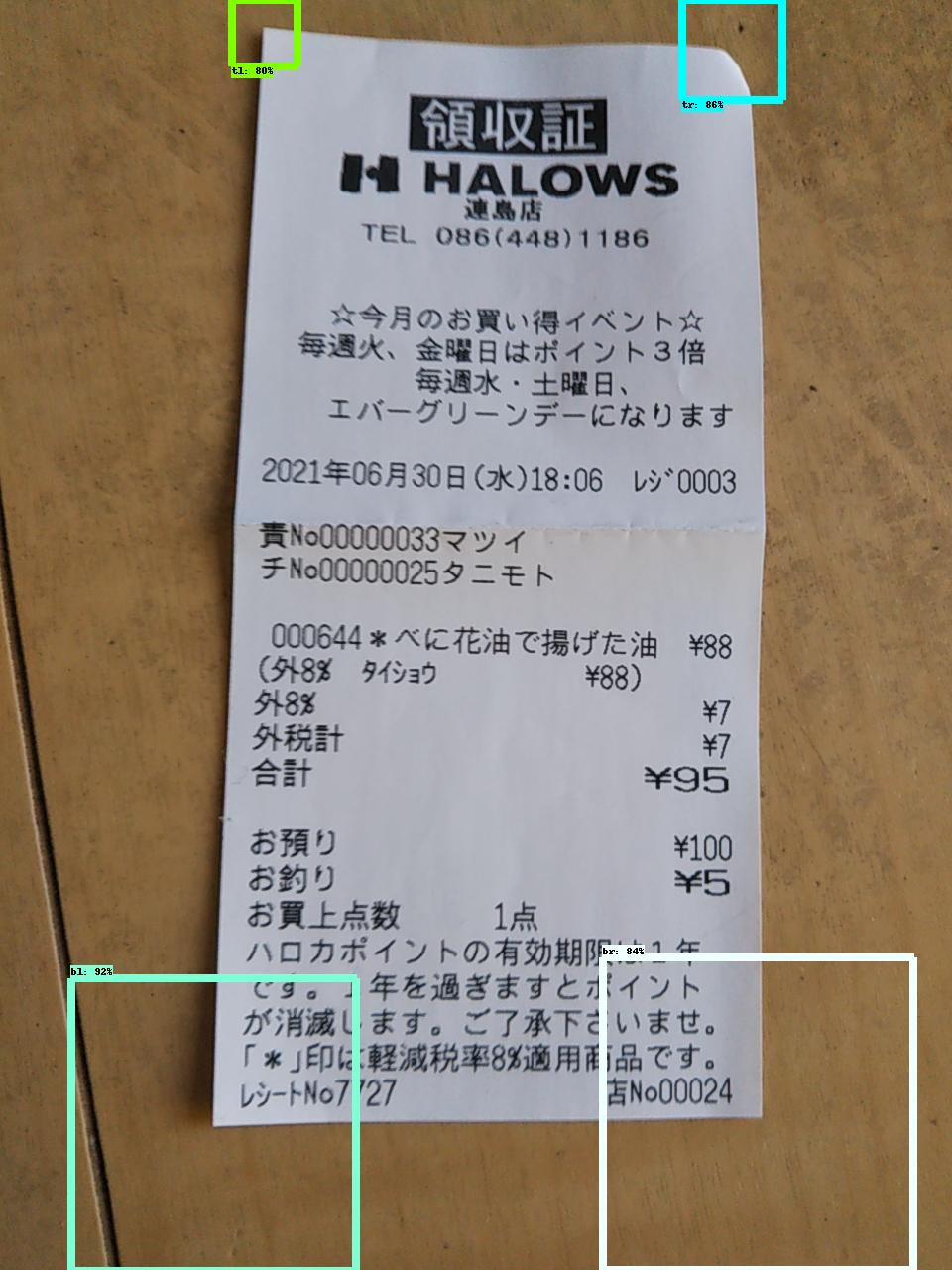}
  \includegraphics[width=2cm]{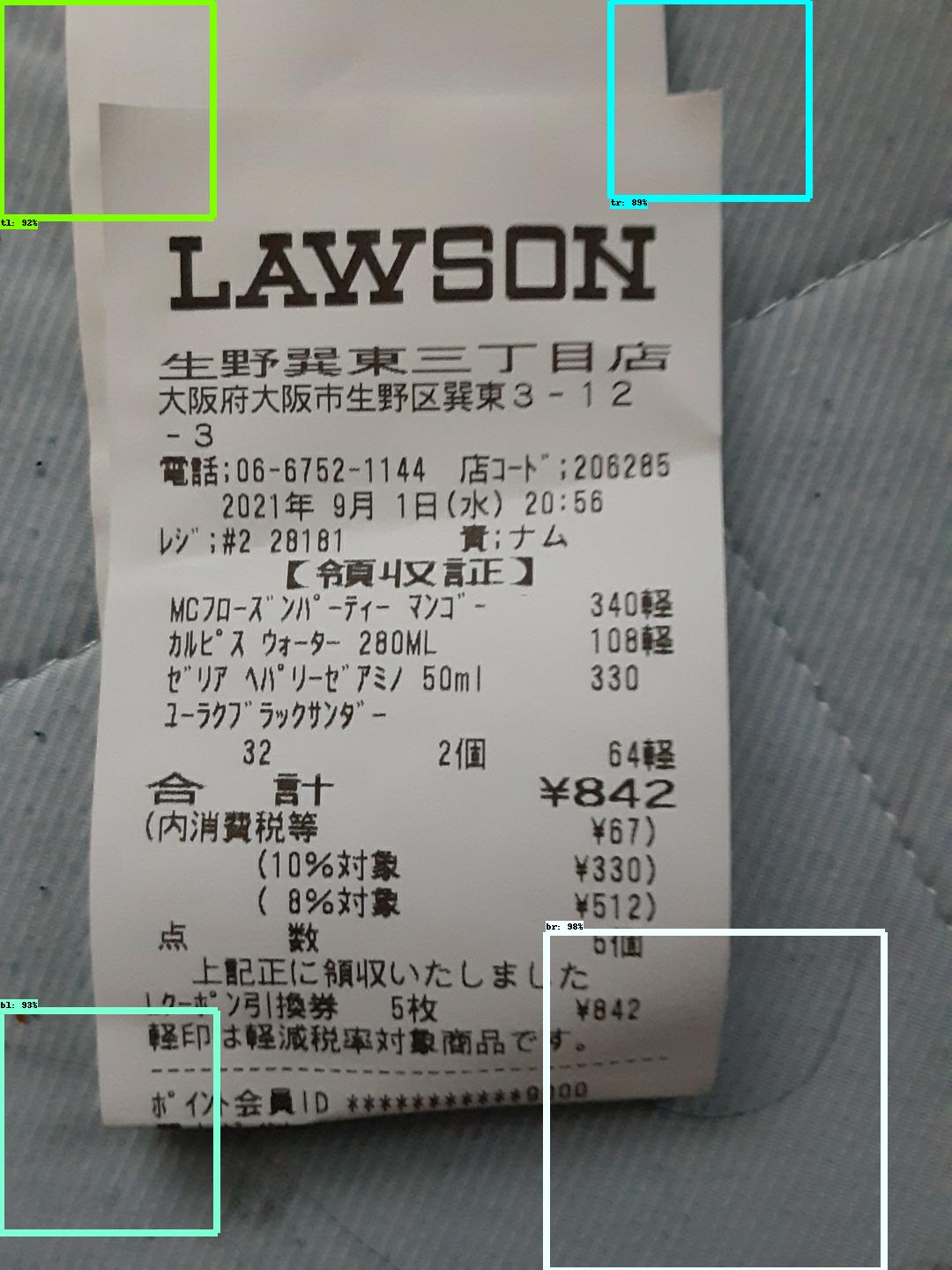}
  \includegraphics[width=2cm]{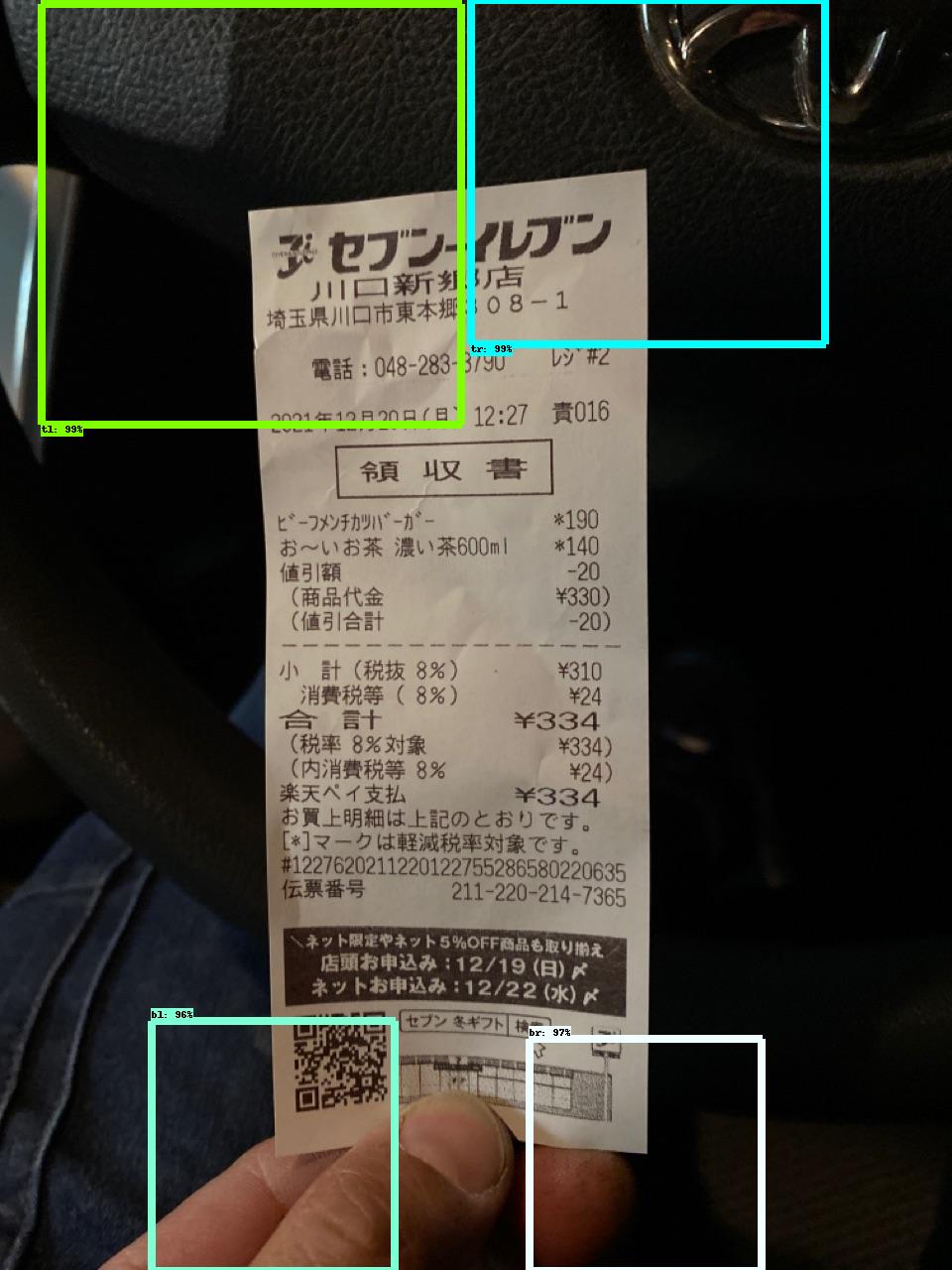}
  \includegraphics[width=2cm]{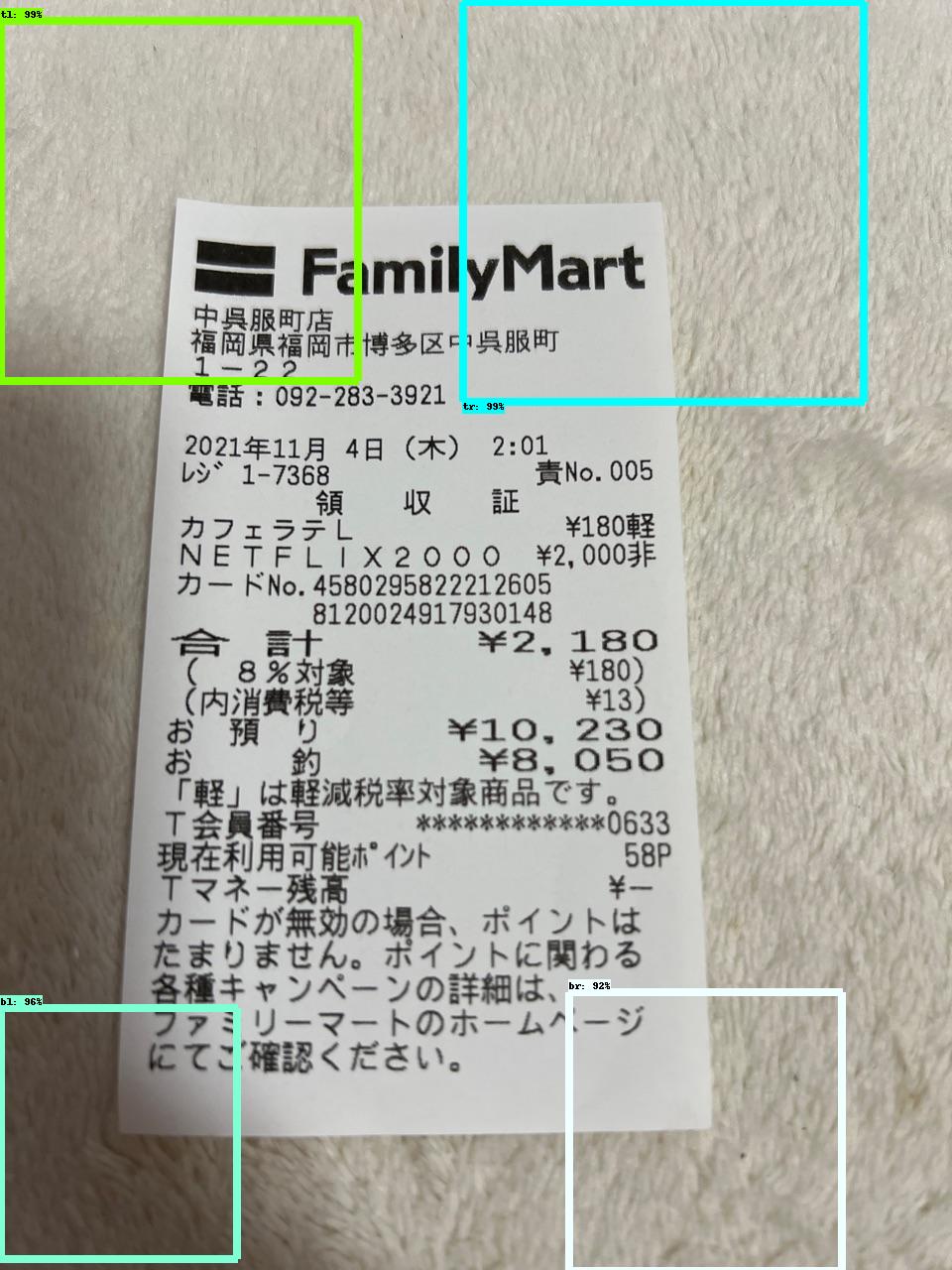}
  \includegraphics[width=2cm]{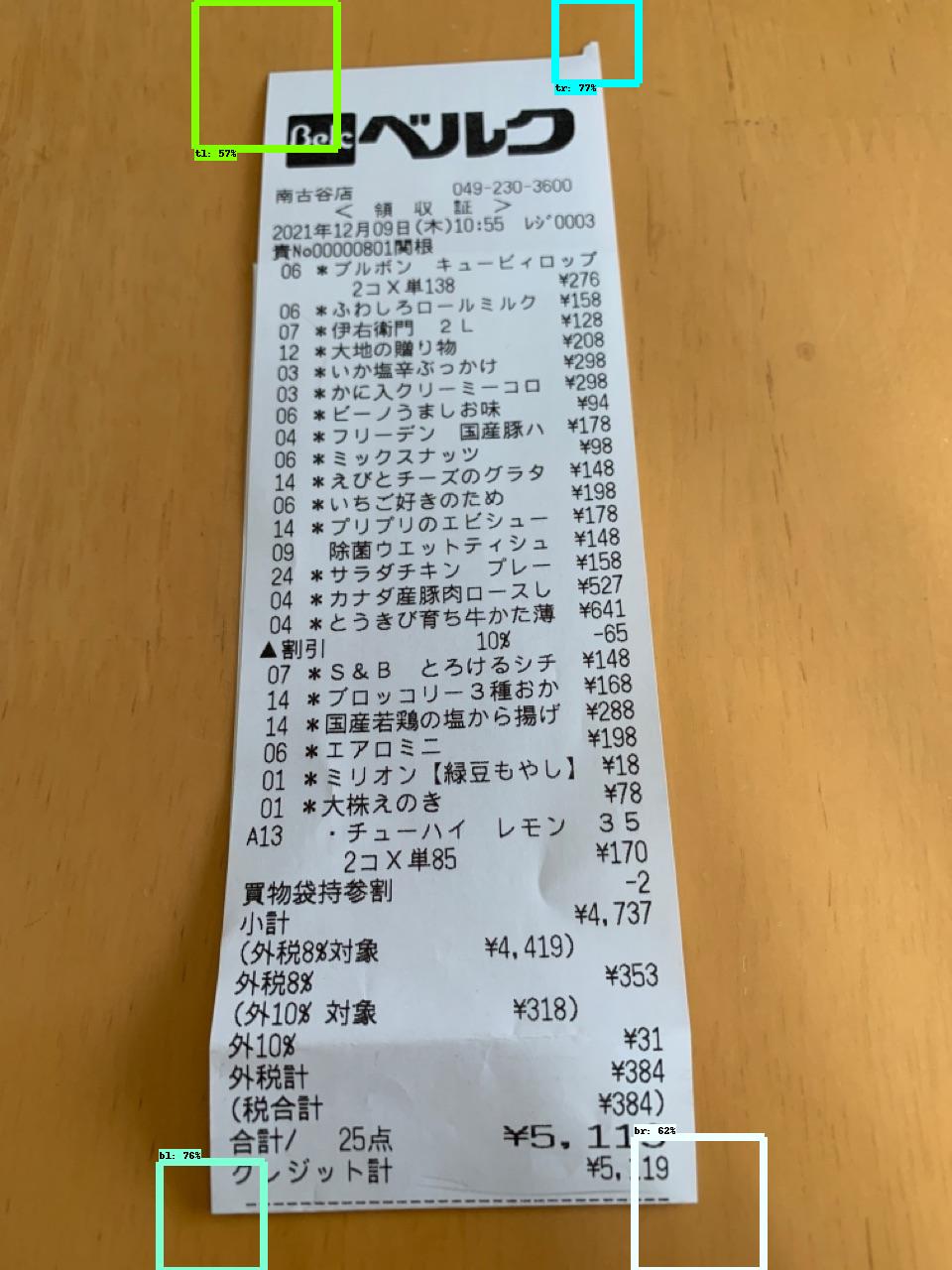}
  \includegraphics[width=2cm]{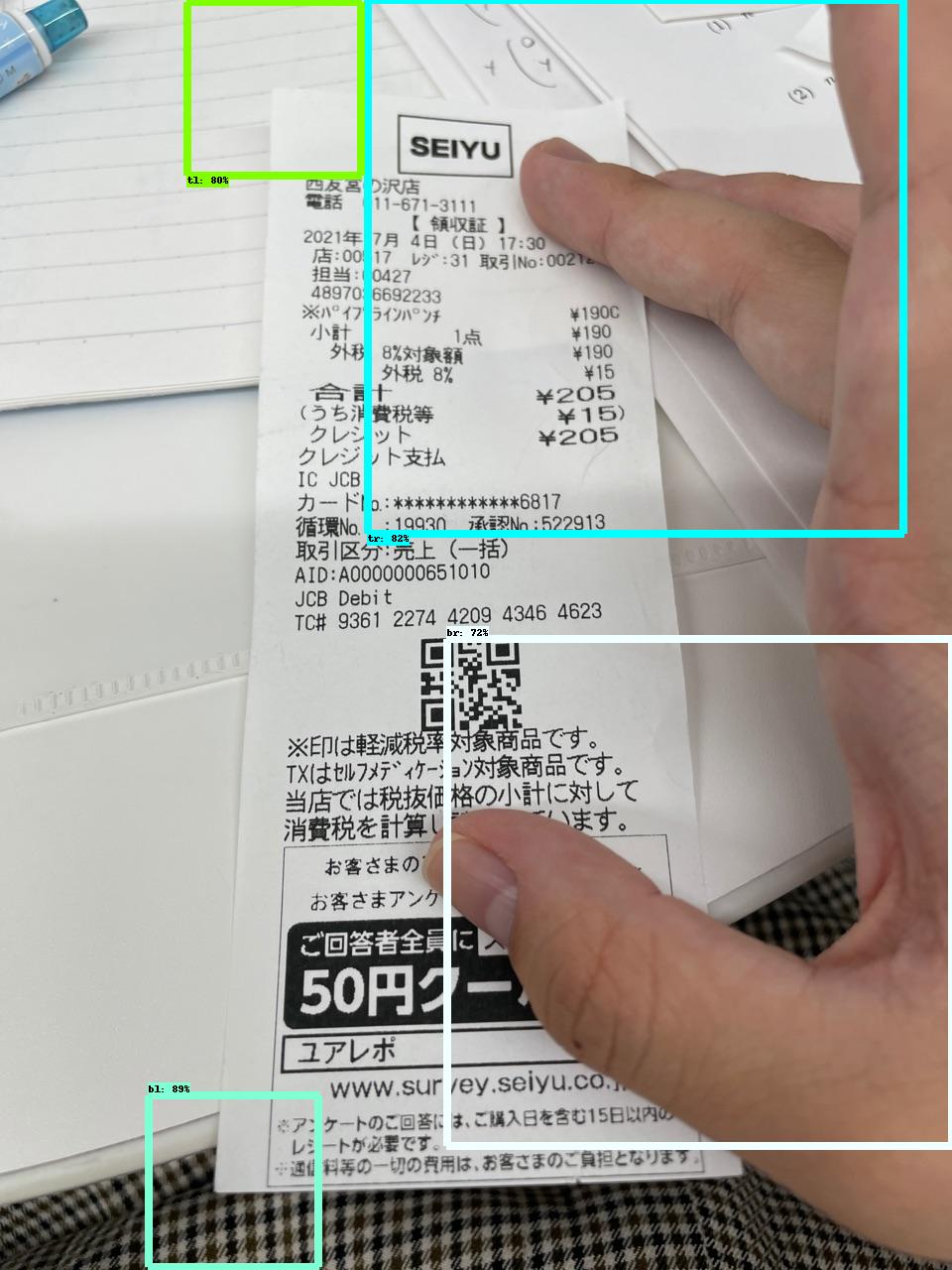}
  \caption{The eight example receipt images from Figure~\ref{ReceiptExamples} showing the detected corner bounding boxes from performing object detection using the proposed method trained on a combination of real and synthetic data.}
  \Description{}
  \label{ReceiptExamplesAfterDetection}
\end{figure}

\begin{figure}[h]
  \centering
  \includegraphics[height=3.5cm]{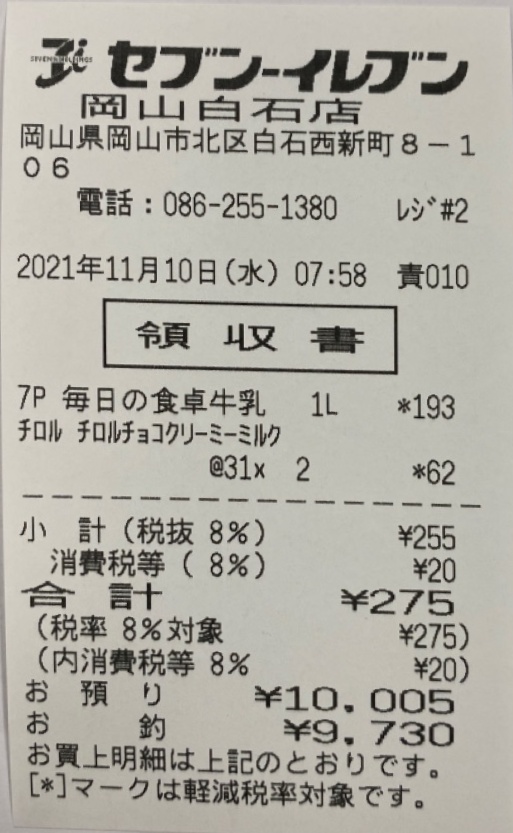}
  \includegraphics[height=3.5cm]{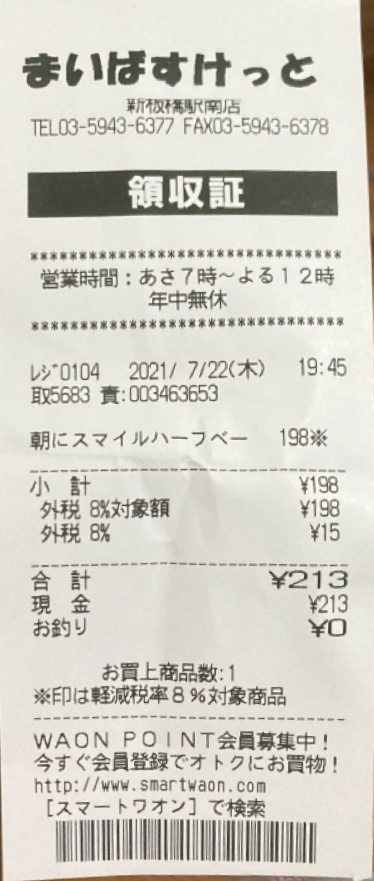}
  \includegraphics[height=3.5cm]{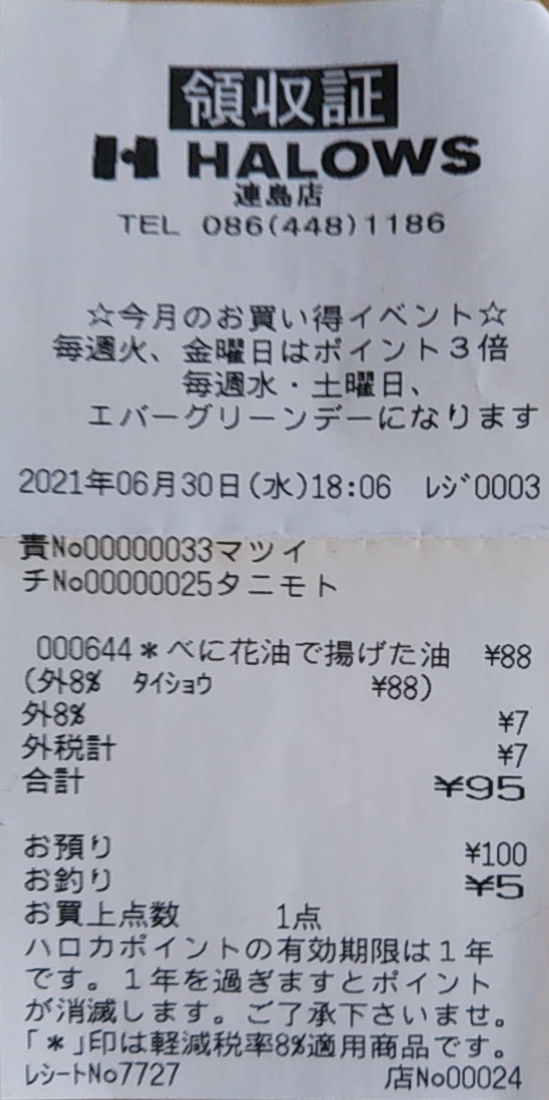}
  \includegraphics[height=3.5cm]{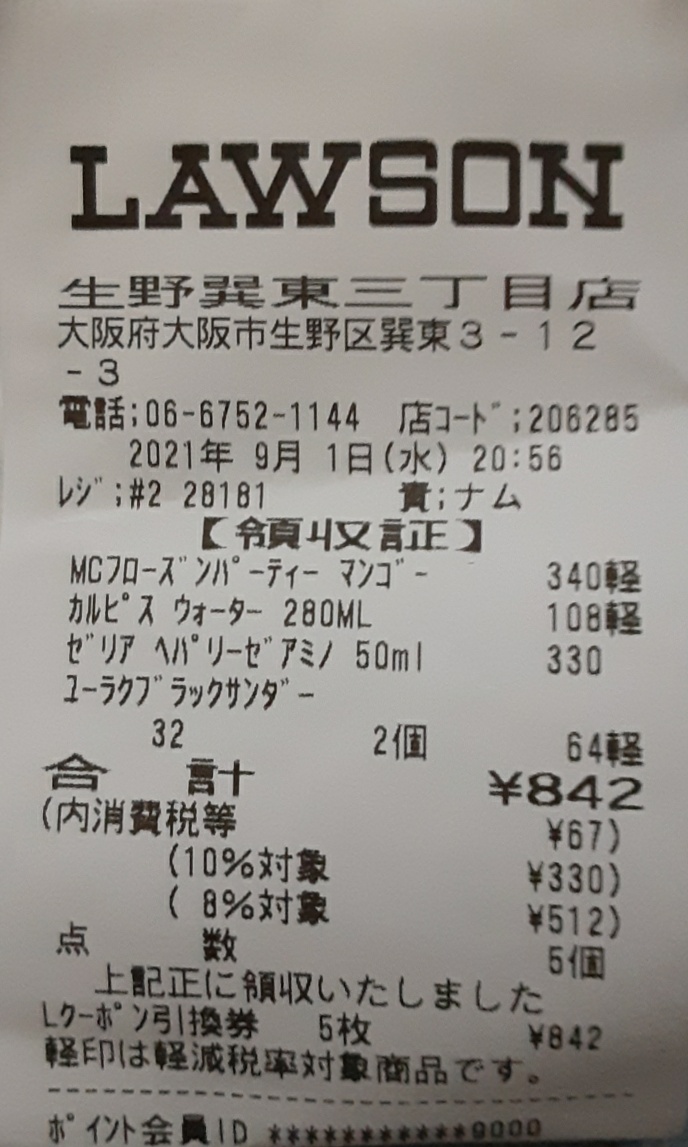}
  \includegraphics[height=3.7cm]{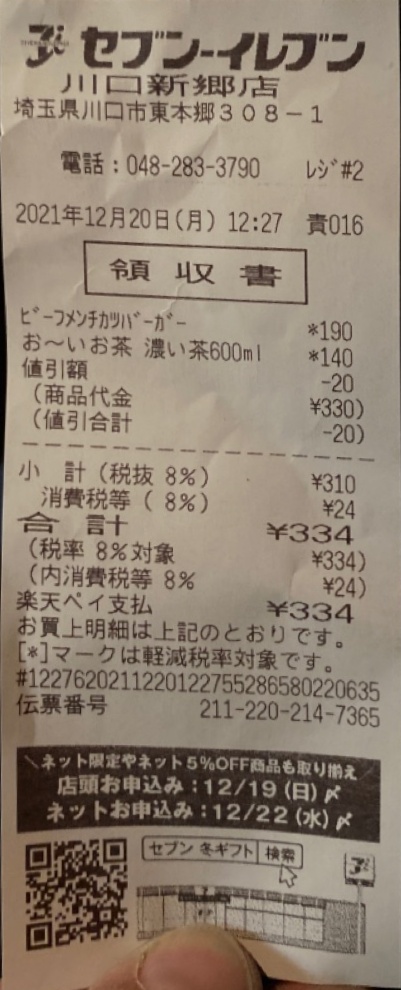}
  \includegraphics[height=3.7cm]{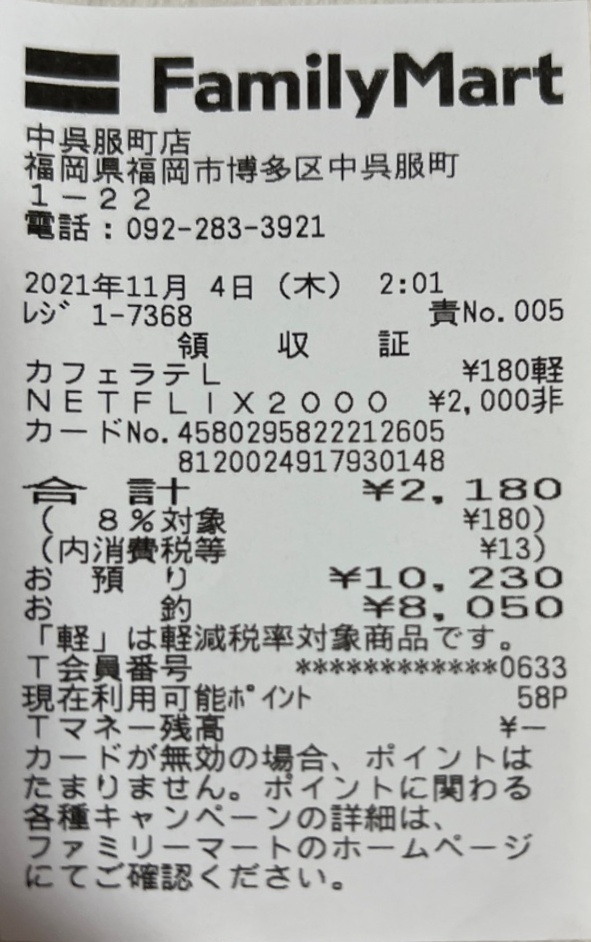}
  \includegraphics[height=3.7cm]{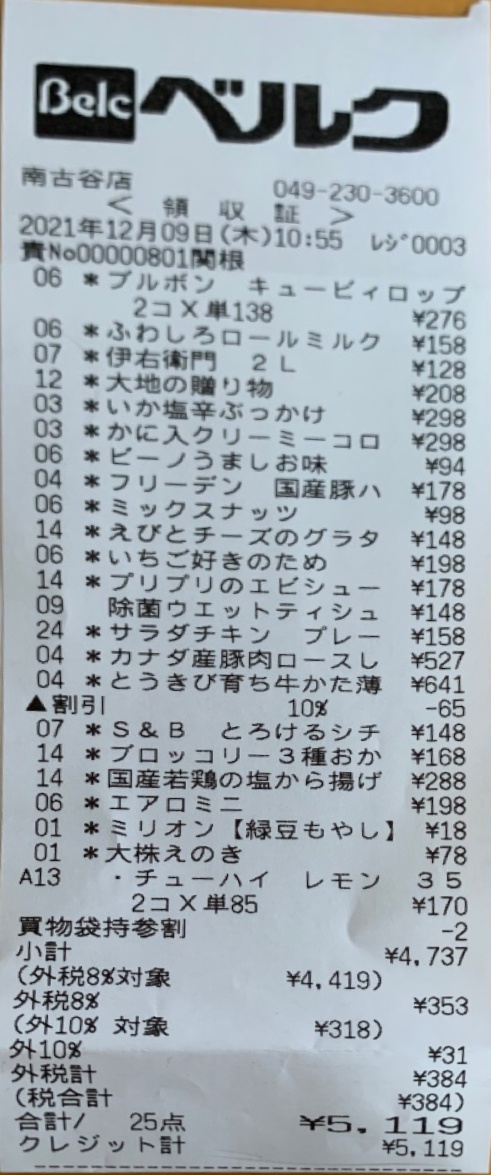}
  \includegraphics[height=3.7cm]{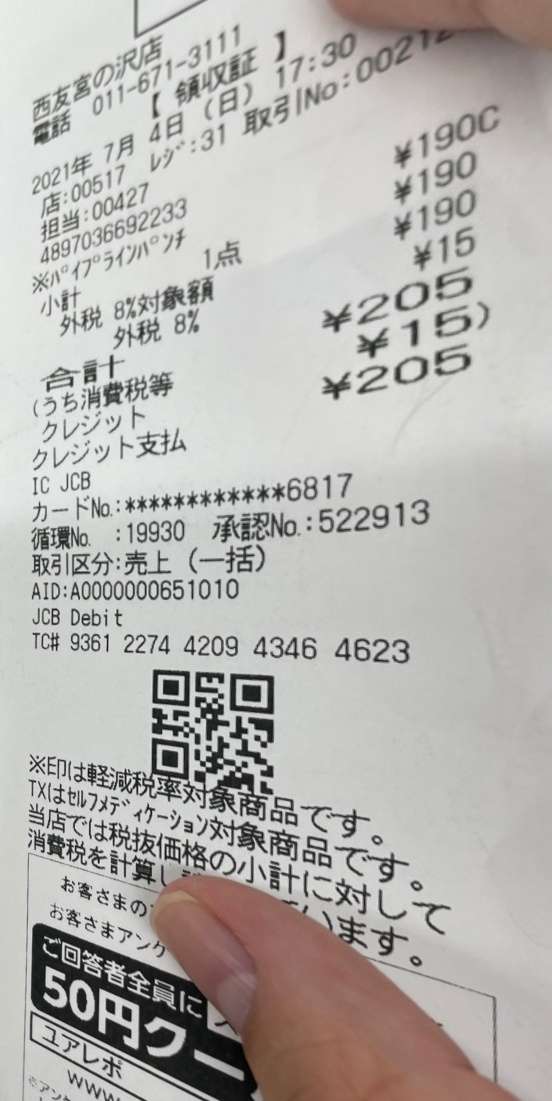}
  \caption{The eight example receipt images from Figure~\ref{ReceiptExamples} showing the automatically extracted and rectified images using the corners shown in Figure~\ref{ReceiptExamplesAfterDetection}.}
  \Description{}
  \label{ReceiptExamplesAfterExtraction}
\end{figure}

In Figure~\ref{ReceiptExamplesAfterDetection} we show the bounding boxes detected on the receipts shown in Figure~\ref{ReceiptExamples} using our proposed method with a model trained using a combination of real and synthetic data. In Figure~\ref{ReceiptExamplesAfterExtraction} we show the result of extraction and rectification using the same detected corners. We observe that most of the receipt corners have been detected correctly and that the resulting rectified receipt images are rectangular and appear to be taken from a ``bird's-eye-view''. The notable exception is the receipt at the bottom-right which is very distorted. We surmise that the model was unable to correctly detect the \texttt{tr} (top-right) and \texttt{br} (bottom-right) receipt corners due to the proximity of the fingers occluding salient image information. Introducing noise and image dropout during training might improve robustness to such phenomenon.

In our proposed smartphone application, instead of using only a single image, as in these experiments, receipt acquisition would typically proceed in a continuous manner using a constant stream of input images, as in a movie recording rather than the single static images that we have been describing. So, in reality we would potentially have a much greater opportunity to select high quality images (e.g. un-blurred and un-obscured receipts) and only use images where all four corners have been detected with high confidence. We can also estimate the blurriness of an image and discard it if it exceeds some threshold. 

In this way, the effort of the user can be significantly reduced, allowing a more relaxed position from which to scan receipts and removing the need to align the physical receipt within the viewfinder or cross-hairs.

\section{Conclusions}\label{Conclusions}

We have described a novel method using object detection for the detection, extraction and rectification of paper receipts on low-contrast backgrounds in the presence of interfering objects.

We have demonstrated the benefits of our proposed approach which, when compared to a traditional computer vision approach using edge detection,
exhibits much higher corner detection accuracy, better discrimination of irrelevant corners and edges in the image, and robustness to variation in the colour and texture of receipts and backgrounds.

Our best proposed method, which was trained on a combination of real and synthetic image data, correctly recognised all four receipt corners in 85.3\% of cases, compared to only 36.9\% using the baseline method. Moreover, our method is easily trainable and is expected to improve further by training on more, and more varied, real receipt images. Such scope for further data-driven improvement is not possible with the baseline method.

In future work we will look at identifying receipt corners that are not visible (e.g.~because they are occluded or missing) by inferring their position using the intersection of recognized partial receipt edges, and using multiple consecutive ``movie''-like images from the streaming camera output. We also plan to investigate the automatic rectification of images where the original paper receipt is curved or folded \cite{Ma_2018_CVPR}\cite{DBLP:journals/corr/YouMSBI16}. 

\begin{acks}
The authors would like to thank Dr. Hans Dolfing for his feedback on an earlier version of this paper.
\end{acks}

\bibliographystyle{ACM-Reference-Format}
\bibliography{Whittaker-Receipt-Rectification-arXiv-Sumbission-2023}

\end{document}